%% file: paper.tex
\PassOptionsToPackage{table}{xcolor}
\documentclass{article}

\usepackage{iclr2027_conference,times}

\usepackage[utf8]{inputenc}
\usepackage[T1]{fontenc}
\usepackage{amsmath,amssymb}
\usepackage{graphicx}
\usepackage{xcolor}
\usepackage{booktabs}
\usepackage{xspace}
\usepackage[inline]{enumitem}
\usepackage{tikz}
\usetikzlibrary{arrows.meta, positioning, calc, shapes.geometric}
\usepackage{subcaption}
\usepackage{caption}
\usepackage{float}
\usepackage{listings}
\usepackage[normalem]{ulem}
\usepackage{upgreek}
\usepackage[bottom]{footmisc}
\usepackage{dirtytalk}
\usepackage{makecell}
\usepackage{pifont}

\newif\ifhassiunitx
\IfFileExists{siunitx.sty}{\hassiunitxtrue}{\hassiunitxfalse}
\ifhassiunitx
  \usepackage{siunitx}
\else
  \newcommand{\qty}[2]{#1\,#2}
  \newcommand{\num}[1]{#1}
  \newcommand{\percent}{\%}
  \newcommand{\second}{s}
\fi
\usepackage{acro}

\DeclareAcronym{LLM}{
  short = LLM,
  long  = Large Language Model
}
\DeclareAcronym{IMO}{
  short = IMO,
  long  = International Mathematical Olympiad
}
\DeclareAcronym{IG}{
  short = IG,
  long  = Inferred Goal
}
\DeclareAcronym{SDNF}{
  short = $S_{\mathrm{DNF}}$,
  long  = Declarative Normal Form Statement
}
\DeclareAcronym{SOTA}{
  short = SOTA,
  long  = state-of-the-art
}

\definecolor{keywordcolor}{rgb}{0.7, 0.1, 0.1}
\definecolor{commentcolor}{rgb}{0.4, 0.4, 0.4}
\definecolor{symbolcolor}{rgb}{0.0, 0.1, 0.6}
\definecolor{sortcolor}{rgb}{0.1, 0.5, 0.1}
\definecolor{errorcolor}{rgb}{1, 0, 0}
\definecolor{stringcolor}{rgb}{0.5, 0.3, 0.2}

\lstdefinestyle{tightlean}{
  language=lean,
  basicstyle=\ttfamily\scriptsize,
  breaklines=true,
  columns=fullflexible,
  keepspaces=true,
  showstringspaces=false,
  aboveskip=2.5pt,
  belowskip=2.5pt,
  xleftmargin=1em,
}
\lstdefinestyle{leanmath}{
  language=lean,
  mathescape=true,
  basicstyle=\ttfamily\scriptsize,
  breaklines=true,
  columns=fullflexible,
  keepspaces=true,
  showstringspaces=false,
  aboveskip=2.5pt,
  belowskip=2.5pt,
  xleftmargin=1em,
  literate={//}{{$\mathbin{/\mkern-6mu/}$}}2,
}

\definecolor{leancodebg}{gray}{0.93}
\newcommand{\lean}[1]{\begingroup\setlength{\fboxsep}{1.5pt}\colorbox{leancodebg}{\texttt{\footnotesize #1}}\endgroup}
\newcommand{\newparagraph}[1]{\textbf{#1.}\,}

\newcommand{\scaletable}[2][\tablescale]{\scalebox{#1}{#2}}

\usepackage[colorlinks=true, allcolors=blue]{hyperref}
\usepackage{cleveref}
\crefname{appendix}{appendix}{appendices}
\Crefname{appendix}{Appendix}{Appendices}

\definecolor{revblue}{RGB}{0, 0, 0}
\newcommand{\rev}[1]{\textcolor{revblue}{#1}}
\newenvironment{revblock}{\color{revblue}}{}
\lstdefinestyle{revlean}{
  language=lean,
  basicstyle=\ttfamily\scriptsize\color{revblue},
  identifierstyle=\ttfamily\color{revblue},
  keywordstyle=[1]\ttfamily\bfseries\color{revblue},
  keywordstyle=[2]\ttfamily\bfseries\color{revblue},
  keywordstyle=[3]\ttfamily\bfseries\color{revblue},
  commentstyle=\ttfamily\itshape\color{revblue!75!black},
  stringstyle=\ttfamily\color{revblue},
  frame=leftline,
  framerule=1.5pt,
  rulecolor=\color{revblue},
  breaklines=true,
  columns=fullflexible,
  keepspaces=true,
  showstringspaces=false,
  aboveskip=3pt,
  belowskip=3pt,
  xleftmargin=1em,
}

\newcommand{\SNL}{\ensuremath{S_{\mathrm{NL}}}\xspace}
\newcommand{\SF}{\ensuremath{S_{\mathrm{F}}}\xspace}

\newcommand{\SDNF}{\ensuremath{S_{\mathrm{DNF}}}\xspace}

\newcommand{\PF}{\ensuremath{P_{\mathrm{F}}}\xspace}
\newcommand{\PNL}{\ensuremath{P_{\mathrm{NL}}}\xspace}

\newcommand{\CF}{\ensuremath{C_{\mathrm{F}}}\xspace}
\newcommand{\CNL}{\ensuremath{C_{\mathrm{NL}}}\xspace}

\newcommand{\modelQwenLong}{Qwen3.6-27B\xspace}
\newcommand{\modelGoedelLong}{Goedel-Formalizer-V2-32B\xspace}

\newcommand{\modelGemmaLong}{Gemma4-31B\xspace}

\newcommand{\modelQwenThree}{Qwen3-32B\xspace}
\newcommand{\modelQwenThreeEight}{Qwen3.8-27B\xspace}

\newcommand{\datasetOmni}{\textsc{Omni-MATH}\xspace}
\newcommand{\datasetImo}{\textsc{IMO-Unformalized}\xspace}
\newcommand{\datasetImoFormalized}{\textsc{IMO-Formalized}\xspace}
\newcommand{\datasetOmniN}{300}
\newcommand{\datasetImoN}{175}
\newcommand{\datasetImoFormalizedN}{223}

\newcommand{\methodOurs}{\textsf{Sage}\xspace}
\newcommand{\methodOursLong}{Semantic Agent-Guided Formalization Engine\xspace}
\newcommand{\methodMonolithic}{\textsf{Monolithic}\xspace}
\newcommand{\methodDecomposition}{\textsf{Decomposed}\xspace}
\newcommand{\methodGoedel}{\textsf{Goedel}\xspace}
\newcommand{\methodGoedelLong}{\textsf{Goedel-Formalizer-V2}\xspace}

\newcommand{\methodSyntaxLoop}{\textsf{Decomposed}+\textsf{Syntax}\xspace}
\newcommand{\methodMonoSyntaxLoop}{\textsf{Monolithic}+\textsf{Syntax}\xspace}
\newcommand{\methodSyntaxLoopAblation}{+\textsf{Syntax}\xspace}

\newcommand{\methodSemanticLoop}{\textsf{Decomposed}+\textsf{Dual}\xspace}
\newcommand{\methodMonoSemanticLoop}{\textsf{Monolithic}+\textsf{Dual}\xspace}
\newcommand{\methodDualLoopAblation}{+\textsf{Dual}\xspace}

\newcommand{\metricCompile}{\textsf{Compile}\xspace}
\newcommand{\metricPWR}{\textsf{PWR}\xspace}

\newcommand{\metricOursSM}{\textsf{Sage-SM}\xspace}

\newcommand{\metricCompileAndOursSM}{\metricCompile \ensuremath{\wedge} \metricOursSM}

\newcommand{\metricGoedelSM}{\textsf{Goedel-SM}\xspace}

\newcommand{\metricCompileAndGoedelSM}{\metricCompile \ensuremath{\wedge} \metricGoedelSM}

\title{Sage:\\ Formalization with Semantic Correction}

\author{Thomas Hirtz\thanks{Equal contribution.} \And
	Farzad Jafarrahmani\footnotemark[1] \And
	Abdelmouksit Sagueni \And
	Xiang Zhou \And
	Wenping Deng \And
	Liang Zhang \AND
	\normalfont Huawei Lagrange Mathematics and Computing Research Center \\
	\normalfont Paris, France
}

\iclrfinalcopy

\begin{document}
\maketitle
\ificlrfinal
\lhead{ArXiv Preprint}
\fi

\begin{abstract}
	While neural theorem provers have achieved impressive milestones in formal mathematics, they largely operate on the assumption that faithful formal statements are already provided.
	Translating informal natural language into a formal language is a critical data bottleneck plagued by an ``illusion of rigor'': standard type-checkers accept statements that compile but drop hypotheses, introduce vacuous truths, or subtly alter mathematical bounds.
	To resolve this, we introduce \methodOurs{} (\methodOursLong{}), an agentic framework that replaces monolithic translation with a four-stage decomposed generation pipeline coupled with a dual-signal semantic correction loop.
	By pairing formal compiler diagnostics with multi-dimensional semantic feedback, our correction loop enforces mathematical fidelity alongside syntactic validity.
	By explicitly accounting for the gap between open-ended queries and declarative formal targets, our pipeline prevents models from achieving high formalization rates by guessing unverified answers (exhibiting a \qty{70.9}{\percent} answer leakage rate in monolithic baselines). Consequently, \methodOurs{} suppresses leakage to \qty{2.7}{\percent} while achieving \textbf{\qty{73.3}{\percent}} pass@4 joint compilation and semantic fidelity on the \datasetOmni{} benchmark without proofs (compared to \qty{42.0}{\percent} for a fine-tuned \methodGoedelLong{} baseline).
	Finally, on \datasetImo{}, a novel frontier of \datasetImoN{} unformalized International Mathematical Olympiad problems, \methodOurs{} demonstrates effective zero-shot generalization with \textbf{\qty{87.4}{\percent}} pass@4 verified fidelity compared to just \textbf{\qty{19.4}{\percent}} for the baseline, winning over \textbf{\qty{79}{\percent}} of blind pairwise evaluations.
\end{abstract}

\section{Introduction}
\label{sec:intro}

\begin{figure}[!t]
	\centering
	\scalebox{0.95}{
		\begin{subfigure}[b]{0.31\textwidth}
			\centering
			\begin{minipage}[c][3.8cm][c]{\linewidth}
				\centering
				\resizebox{0.90\textwidth}{!}{%
				\begin{tikzpicture}[
					font=\small,
					box/.style={draw=black!60, rounded corners=3pt, align=left,
						inner sep=5pt, fill=white},
					arr/.style={-{Latex}, thick, draw=black!70}
					]
					\node[box, text width=4.7cm] (nl) {
						\textbf{Informal}~\SNL\\[3pt]
						\emph{There are infinitely many primes.}};
					\node[box, text width=4.7cm, below=0.65cm of nl] (sf) {
						\textbf{Formal}~\SF\\[3pt]
						{\ttfamily\scriptsize
						\textcolor{keywordcolor}{theorem} infinitely\_many\_primes :\\
						\quad $\forall$\,N,\,$\exists$\,p\,$\geq$\,N,\\
						\quad Nat.Prime\,p\,\textcolor{keywordcolor}{:= by}\,sorry}};
					\draw[arr] (nl.south) -- node[right] {Translation} (sf.north);
				\end{tikzpicture}%
				}
			\end{minipage}
			\caption{Autoformalization example.}
			\label{fig:formalization-example}
		\end{subfigure}
		\hfill
		\begin{subfigure}[b]{0.29\textwidth}
			\centering
			\begin{tikzpicture}[
				font=\scriptsize,
				node distance=0.28cm,
				box/.style={draw=black!60, rounded corners=2pt, align=center,
					minimum height=4.5mm, inner sep=2pt, fill=white},
				input/.style={box, fill=gray!8, dashed},
				agent/.style={box, fill=blue!6},
				terminal/.style={box, fill=gray!12},
				arr/.style={-Latex, thick, draw=black!70}
				]
				\node[input]    (SNL)  {\SNL};
				\node[input, right=0.35cm of SNL] (PNL) {\PNL{} (opt.)};

				\node[agent, below=0.55cm of $(SNL)!0.5!(PNL)$] (dist) {Distiller};
				\node[agent, below=of dist] (prep) {Preprocessor};
				\node[agent, below=of prep] (form) {Formalizer};
				\node[agent, below=of form] (fmt)  {Formatter};
				\node[terminal, below=of fmt] (SF) {\SF};

				\draw[arr] (SNL.south) -- ++(0,-0.1) -| (dist.north);
				\draw[arr] (PNL.south) -- ++(0,-0.1) -| (dist.north);
				\draw[arr] (dist) -- (prep);
				\draw[arr] (prep) -- (form);
				\draw[arr] (form) -- (fmt);
				\draw[arr] (fmt)  -- (SF);
			\end{tikzpicture}
			\caption{Decomposed generation.}
			\label{fig:decomposition}
		\end{subfigure}
		\hfill
		\begin{subfigure}[b]{0.4\textwidth}
			\centering
			\begin{tikzpicture}[
				font=\tiny,
				box/.style={draw=black!60, rounded corners=2pt, align=center,
					minimum height=4.5mm, inner sep=2pt},
				terminal/.style={box, fill=gray!12, font=\scriptsize},
				llm/.style={box, fill=blue!8},
				ver/.style={box, fill=orange!15},
				gate/.style={draw=black!60, diamond, aspect=1.8, align=center, inner sep=0.5pt, fill=yellow!15, font=\tiny},
				arr/.style={-Latex, thick, draw=black!70},
				line/.style={thick, draw=black!70},
				looparr/.style={-Latex, thick, dashed, draw=black!70},
				]
				\node[terminal] (SF) at (0, 0) {\SF};

				\node[ver] (V1) at (-1.2, -1.0) {Syntax\\Verifier};
				\node[llm] (R)  at (1.2, -1.0)  {Semantic\\Rater};
				\node[ver] (V2) at (1.2, -1.95)  {Rater's\\Correction\\Verifier};

				\node[gate] (G)   at (-0.0, -3.4)     {Compile $\wedge$\\Ratings Ok?};
				\node[llm]  (Cor)  at (-2.1, -3.4) {Statement\\Corrector};
				\node[terminal] (Exit) at (1.8, -3.4) {$S_F^\ast$};

				\coordinate (split) at (0, -0.35);
				\draw[line] (SF.south) -- (split);
				\draw[arr] (split) -| (V1.north);
				\draw[arr] (split) -| (R.north);

				\draw[arr] (R.south) -- (V2.north);

				\coordinate (join) at (-0.0, -2.55);
				\draw[line] (V1.south) |- (join);
				\draw[line] (V2.south) |- (join);
				\draw[arr] (join) -- (G.north);

				\draw[arr] (G.west) -- node[midway, above, font=\tiny] {No} (Cor.east);
				\draw[arr] (G.east) -- node[midway, above, font=\tiny] {Yes} (Exit.west);

				\draw[looparr] (Cor.north) |- (SF.west)
				node[pos=0.25, left, align=center, font=\tiny] {Next round\\(Budget $T$)};
			\end{tikzpicture}
			\caption{Correction loop architecture.}
			\label{fig:correction-loop}
		\end{subfigure}
	}
	\caption{
		Overview of \methodOurs{}: an autoformalization example \textbf{(\subref{fig:formalization-example})}, the multi-stage generation chain \textbf{(\subref{fig:decomposition})}, and the syntax/semantic feedback loop \textbf{(\subref{fig:correction-loop})}. Blue: LLM agents; orange: verifiers.
	}
	\label{fig:pipeline}
\end{figure}
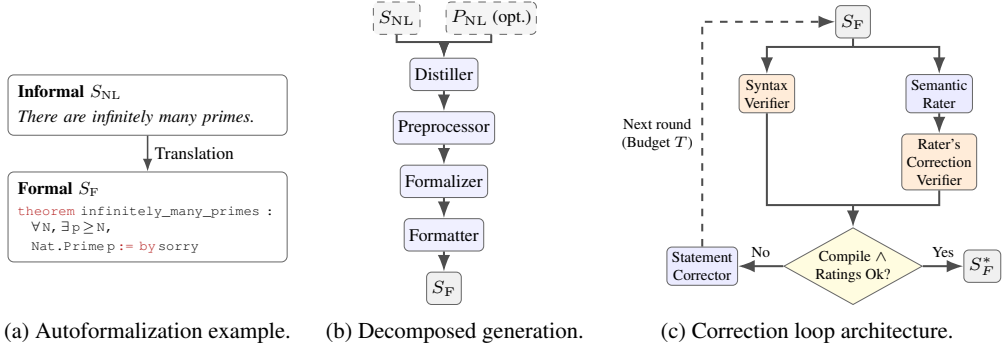

Neural theorem provers and olympiad-scale systems such as Goedel-Prover~\citep{lin2025goedelproverv2}, DeepSeek-Prover~\citep{ren2025deepseekproverv2} and AlphaProof~\citep{hubert2025olympiad} have pushed automated reasoning in Lean~\citep{demoura2021lean4}, yet their benchmarks largely assume the formal statement \SF is already given and focus solely on finding a formal proof ($\SF \to \PF$). In practice, competition problems, textbooks, and user-facing assistants arrive as informal natural language~\SNL; translating $\SNL \to \SF$ (\Cref{fig:formalization-example})
with \emph{semantic fidelity} is the mandatory upstream step and a bottleneck for high-quality formal training data. A faithful~\SF is the contract under which every downstream prover, benchmark, and library entry must operate---yet the field lacks a shared account of what that contract \emph{contains} beyond ``it type-checks.'' Crucially, compile success is not semantic fidelity: Lean~4 accepts statements that compile but fundamentally \emph{misstate} the problem. For example, if a model drops a crucial minimality condition, the resulting bare existential type-checks perfectly with \lean{sorry}. Compilers cannot detect semantic mismatches, leaving standard compile-only repair loops blind to dropped hypotheses, introduced tautologies (\lean{True}), and vacuous placeholders.

\newparagraph{Our approach}
To address this, we introduce \textbf{\methodOurs{}} (\textbf{\methodOursLong{}}), an agentic framework shifting autoformalization from monolithic generation to iterative semantic correction.
\methodOurs{} decomposes formalization into four inspectable stages (distillation, preprocessing, formalization, formatting) and refines candidates via a dual-signal correction loop combining compiler diagnostics with multi-dimensional semantic feedback.
To support distinct downstream applications, \methodOurs{} provides two configurations: \textbf{Answer-Aware} (leveraging informal proofs to curate resolved, prover-ready targets) and \textbf{Answer-Agnostic} (translating problem text alone without proof oracles or solutions). Our contributions are as follows:
\begin{enumerate}[nosep, leftmargin=*]
	\item \textbf{Theoretical Foundations:} We formalize the distinction between Assertion- and Question-type mathematics, exposing \rev{the \emph{Feasibility Trilemma} of closed single-\lean{sorry} translation,} the \emph{Discovery Trap}\rev{,} and type-theoretic vulnerabilities that arise when translating open queries without proof oracles.
	\item \textbf{Decomposed Generation Pipeline:} We introduce a modular four-stage formalization protocol that decouples goal extraction, typing, and syntax generation \rev{into native Mathlib theorems}, isolating structural constraints and eliminating unmonitored answer guessing.
	\item \textbf{\rev{Direct Formal-Language} Semantic Feedback:} We define an actionable rubric tracking critical axes of translation divergence---including dropped hypotheses, altered bounds, and vacuous placeholders---operationalized into a \rev{reference-free} test-time semantic evaluator \rev{acting directly on Lean~4 code without lossy natural-language back-translation}.
	\item \textbf{Dual-Signal Correction Loop:} Our iterative repair loop \rev{simultaneously} pairs Lean~4 compiler diagnostics with \rev{in-turn source-level} semantic feedback \rev{and pre-checked repair drafts}, correcting subtle mathematical flaws that type-checkers overlook.

\end{enumerate}

Empirically, \methodOurs{} outperforms zero-shot generation, syntax-only repair loops, and fine-tuned baselines. On \datasetOmni{} without proofs, \methodOurs{} suppresses unmonitored answer leakage from \qty{70.9}{\percent} to \qty{2.7}{\percent}, while nearly doubling the fine-tuned \methodGoedelLong{} baseline in verified formalization (\textbf{\qty{73.3}{\percent}} vs.\ \qty{42.0}{\percent} pass@4; \textbf{\qty{50.3}{\percent}} vs.\ \qty{26.7}{\percent} pass@1). On the unformalized IMO frontier (\datasetImo{}), \methodOurs{} achieves \textbf{\qty{87.4}{\percent}} pass@4 verified fidelity (vs.\ \qty{19.4}{\percent} for the baseline), winning over \textbf{\qty{79}{\percent}} of blind pairwise head-to-head comparisons.

\Cref{sec:related} situates our work in autoformalization and self-refinement; \Cref{sec:limits} formalizes the theoretical limits of translation;
\Cref{sec:sage,sec:experimental-setup} describe the approach and evaluation design;
\Cref{sec:experiments} reports results; and
\Cref{sec:limitations,sec:conclusion} discuss limitations and conclude.

\section{Related Work}
\label{sec:related}

\newparagraph{Neural Theorem Proving and the Statement Bottleneck}
While automated theorem provers have achieved olympiad-level reasoning~\citep{hubert2025olympiad,lin2025goedelproverv2,ren2025deepseekproverv2,yang2023leandojo}, they operate under the assumption that formal targets \SF are already given. Yet establishing \SF from natural language (\SNL $\to$ \SF) remains an open vulnerability: human annotations frequently encode subtle misinterpretations~\citep{zheng2022minif2f,liu2023fimo} that require retrospective benchmark corrections~\citep{ospanov2025minif2f,poiroux2025reliable}, whereas automated pipelines guided purely by type-checkers tend to compromise mathematical constraints simply to secure compilation~\citep{ammanamanchi2026faults,zhang2026beyond}. Although formalization has also been used downstream to verify multi-step informal reasoning~\citep{zhou2024dont}, our work tackles the primary upstream bottleneck: synthesizing semantically faithful formal targets.

\newparagraph{Autoformalization and Semantic Evaluation}
Autoformalization leverages formal corpora~\citep{azerbayev2024proofnet,ying2024lean} and specialized foundation models~\citep{azerbayev2024llemma}. Yet on open queries without proofs, unguided models frequently drop hypotheses or alter bounds. To evaluate fidelity, \citet{lu2025formalalign} align representation spaces, \citet{liu2025rethinking} introduce bidirectional equivalence (\textbf{BEq}), \citet{han2026shadowbench} test shadow theorems, and \citet{zhang2026beyond} audit compile--faithfulness gaps---metrics that are typically \emph{reference-based} (requiring a human gold $\SF^*$) and thus inapplicable to unformalized frontiers like \datasetImo{}. Roundtrip rerankers~\citep{li2024autoformalize} similarly operate \emph{post-hoc} over static drafts. In contrast, our in-loop gate \metricOursSM{} is \emph{reference-free}, scoring candidates directly against $\SNL$ to drive test-time repair.

\newparagraph{Modular Decomposition and Iterative Refinement}
Modular decomposition~\citep{ridnik2024alphacodium} and iterative self-refinement~\citep{shinn2023reflexion} are well-established paradigms in general code generation.
In formal mathematics, prior systems explore: (i)~compiler-only repair~\citep{lu2024process}; (ii)~monolithic reflective self-critique~\citep{chen2026reform}; or (iii)~dependency-graph retrieval without an in-loop semantic gate against $\SNL$~\citep{wang2025aria}.
\rev{Related workflows either append official answers before monolithic compilation and expert post-editing (\textsf{FIMO};~\citealp{liu2023fimo}), transpile through a closed 40-concept assertional-logic DSL that emits dual-\lean{sorry} templates (\textsf{KELPS};~\citealp{zhang2025kelps}), or generate an in-prompt \lean{MySystem} template in a single call to emit open \lean{def} predicates ($\alpha \to \texttt{Prop}$) refined via segmented natural-language back-translation and alternating loops (\textsf{LoC-Decomp};~\citealp{shi2026locdecomp}).}
To avoid confounding factors from disparate base models and training sets, our $2 \times 3$ ablation matrix (\Cref{sec:experimental-setup}) benchmarks these core mechanisms under an identical backbone and \rev{matched iteration budgets}.
Crucially, compiler-only repair prioritizes syntactic compilation over semantic fidelity, whereas monolithic reflection succumbs to the Discovery Trap on open queries (\Cref{sec:exp-discovery-trap}).
\methodOurs{} resolves both by combining four-stage decomposition with \rev{direct, reference-free Lean~4} semantic gating \rev{(architectural comparison in \Cref{app:prior-frameworks})}.

\section{Formalizing Mathematics: Beyond Naive Translation}

\label{sec:limits}

Autoformalization is a constrained translation task: mapping informal mathematical language (English) into a formal language.
A successful formalization must satisfy the following desiderata:
\begin{enumerate}[nosep, leftmargin=*]
	\item \textbf{Syntactic validity:} the statement must be well-typed in the target system and respect the \emph{single-sorry contract}: $\SF$ must contain strictly one terminating placeholder (\lean{:= by sorry}) closing
    the \emph{primary target declaration} (e.g. \lean{theorem}).

	\item \textbf{Semantic fidelity:} the statement must preserve the mathematical meaning of the source.
	\item \textbf{Texture:} the formalization should respect the pragmatic structure of the source---whether the informal text asserts a declarative claim to be verified, or poses an open-ended question asking for a value, witness, or polarity.
\end{enumerate}
We call the second dimension \emph{semantic match}. In \Cref{sec:method-loop}, \metricOursSM serves as an in-loop fidelity metric to gate our correction loop, and we adopt the existing \metricGoedelSM protocol as a post-hoc judge.

\newparagraph{Assertion-Type and Question-Type Statements: The Discovery vs. Verification Gap}
\rev{Building on prior autoformalization benchmarks that contrast proof-type claims with answer-seeking problems~\citep{zheng2022minif2f,liu2023fimo,zhang2025kelps,shi2026locdecomp}, a} useful distinction for autoformalization is between \emph{Assertion-type} and \emph{Question-type} statements.
An \textbf{Assertion-type statement} already presents a complete mathematical claim to be verified (e.g., ``Show that\dots'' or ``Prove that\dots'').
By contrast, a \textbf{Question-type statement} asks for an unknown object, value, witness, or characterization (e.g., ``Find\dots'', ``Determine\dots'').

This exposes the \emph{Discovery vs. Verification Gap}: natural language math poses open-ended \emph{discovery} problems, whereas theorem provers like Lean~4 require strict \emph{verification}. Formalizing open queries into \lean{theorem} statements forces pre-guessing answers, rendering the simultaneous satisfaction of all three translation dimensions often impossible. Consequently, for question-type statements, preserving original \emph{texture} and ensuring \emph{semantic fidelity} work at cross-purposes, exposing how verification systems fundamentally struggle with mathematical discovery.

\newparagraph{The Feasibility Trilemma}
To bridge this gap and convert open questions into declarative targets, we identify three distinct formalization strategies, each presenting a different trade-off between semantic match and downstream prover usability. Canonical encodings for an abstract query are contrasted in \Cref{tab:feasibility-trilemma}:
\begin{enumerate}[nosep, leftmargin=*]
	\item \textbf{Explicit Answer Injection:} Incorporates the resolved answer directly into the statement, rewriting the open question as a concrete assertion.
	\begin{itemize}[nosep, leftmargin=*]
		\item \textit{Pros:} Produces clean, fully specified declarative theorems ready for downstream tactic provers.
		\item \textit{Cons:} Requires an oracle or proof context to resolve the value prior to formalization, completely breaking the open-ended texture of the original query.
	\end{itemize}
	\item \textbf{Placeholder Definitions:} Introduces a typed, unresolved constant via a placeholder stub, asserting the desired properties about that constant in the theorem.
	\begin{itemize}[nosep, leftmargin=*]
		\item \textit{Pros:} Formalizable without prior knowledge of the answer while preserving original texture.
		\item \textit{Cons:} Forces proving agents to modify the source code (instantiating the placeholder), violating the read-only contract of automated theorem proving.
	\end{itemize}
	\item \textbf{Existential Encoding:} Reformulates the query as an existential proposition over the constraints, delegating constructive witness search to the downstream prover.
	\begin{itemize}[nosep, leftmargin=*]
		\item \textit{Pros:} Faithfully captures open-ended natural language intent without an answer oracle.
		\item \textit{Cons:} Highly fragile when applied to polar queries and susceptible to computational bypass in characterization queries (the \emph{Triviality Trap}; see \Cref{sec:limits-vulnerabilities}).
	\end{itemize}
\end{enumerate}

We unify this fundamental trade-off: an answer-agnostic autoformalizer cannot map an open interrogative query into a single declarative proposition without assuming an oracle, breaking prover usability, or introducing logical vulnerabilities.

\begin{table}[ht]
	\centering
	\caption{\textbf{The Formalization Feasibility Trilemma across Question Types.}}
	\label{tab:feasibility-trilemma}
	\small
	\resizebox{\textwidth}{!}{
		\begin{tabular}{@{}l l cccc@{}}
			\toprule
			\textbf{Strategy}
			& \makecell[l]{\textbf{Lean~4 Formalization Example}\\\footnotesize (\SNL: Find an object $x$ of type $\alpha$ that satisfies the property $P$)}
			& \textbf{\makecell{Syntactic\\Validity?}}
			& \textbf{\makecell{Semantic\\Fidelity?}}
			& \textbf{\makecell{Preserves\\Texture?}}
			& \textbf{\makecell{Requires\\Oracle?}} \\
			\midrule
			\textbf{Explicit Answer}
			& \lean{theorem target : P c := by sorry}
			& \textbf{Yes}
			& \textbf{Yes}
			& No
			& \textbf{Yes} \\
			\addlinespace[3pt]
			\textbf{Placeholder}
			& \makecell[l]{\lean{def x : $\alpha$ := sorry}\\\lean{theorem target : P x := by sorry}}
			& \makecell{Breaks\\Read-Only}
			& \textbf{Yes}
			& \textbf{Yes}
			& No \\
			\addlinespace[3pt]
			\textbf{Existential}
			& \lean{theorem target : $\exists$ x : $\alpha$, P x := by sorry}
			& \textbf{Yes}
			& Vulnerable
			& \textbf{Yes}
			& No \\
			\bottomrule
		\end{tabular}%
	}
\end{table}

Crucially, reducing open interrogative queries to standard \lean{Prop} existentials allows automated provers to bypass constructive search via syntactic shortcuts, such as the \emph{Polarity Cheat} and the \emph{Triviality Trap}.
To prevent this, we propose \emph{Type-Theoretic Locks} to structurally enforce witness synthesis.
Within \methodOurs{}, our Formalizer and in-loop Rater actively apply lock-aware rules to steer candidate statements away from trivialized encodings: where feasible, the pipeline attempts to harden statements into locked formulations, otherwise defaulting to a standard existential.
We detail the underlying mechanics, criteria, and reference Lean~4 lock implementations in \Cref{app:type_locks}.

Rather than treating this trilemma as an insurmountable barrier, \methodOurs{} deliberately navigates it by supporting two distinct operating configurations. In the \textbf{Answer-Agnostic (Fidelity-Oriented)} regime, we adopt \emph{Existential Encodings} to strictly preserve query texture and prevent answer leakage. Conversely, in the \textbf{Answer-Aware (Prover-Oriented)} regime, we utilize \emph{Explicit Answer Injection} guided by an informal proof, providing automated tactic engines with the fully resolved, concrete goals they require.

\section{The \methodOurs{} Framework: Composition \& Semantic Correction}
\label{sec:sage}

To prevent the silent semantic failures of single-pass translation, \methodOurs{} decomposes autoformalization into three sequential phases: \textbf{Draft} (compositional generation), \textbf{Verify} (multi-dimensional syntactic and semantic auditing), and \textbf{Refine} (dual-signal iterative correction).

\subsection{Compositional Generation (The Draft Phase)}
\label{sec:sage-generation}

Given an informal statement $\SNL$ and an optional informal proof $\PNL$, generation maps $(\SNL, \PNL?) \to \SF$ through four specialized, role-isolated agents (\Cref{fig:decomposition}):
\begin{enumerate}[nosep, leftmargin=*]
	\item \textbf{Distiller:} Evaluates $\SNL$ and informal proof $\PNL$ (when available). In the answer-aware regime, it classifies the problem type (Assertion- vs. Question-type) and extracts the target witness or characterization from the proof as an \ac{IG}; in the answer-agnostic regime, it reformulates open interrogative queries into explicit declarative targets without an answer oracle.
	\item \textbf{Preprocessor:} Standardizes $\SNL$ via information provided by the distiller into a pseudocode representation, isolating supporting mathematical definitions, localized hypotheses ($h_1, \dots, h_n$), and the target conclusion ($G$).
	\item \textbf{Formalizer:} Translates the preprocessed pseudocode into idiomatic Lean~4 declarations and theorem signatures, leveraging standardized \textit{Mathlib} abstractions \citep{mathlib2020} while avoiding trivialization stubs.
	\item \textbf{Formatter:} Performs syntactic sanitization and assembly, organizing imports, namespaces, and \lean{noncomputable} sections to ensure the draft is a standalone, compilable Lean~4 file without altering its mathematical semantics.
\end{enumerate}
We provide the details of each agent in \Cref{app:decomposition-details} as well as a step-by-step walkthrough across both regimes on \datasetOmni{} \#043 in \Cref{app:walkthrough} (\Cref{fig:pipeline-example}).

\newparagraph{Syntax Verification and the Single-Sorry Contract}
An external Lean~4 verifier type-checks each candidate statement \SF (\metricCompile), deferring proof bodies via \lean{:= by sorry} as our goal is statement formalization rather than proof search.
When compilation fails, structured diagnostics (location and error messages) are routed directly to the correction loop (\Cref{sec:method-loop}).
Crucially, a prover-ready formalization must also satisfy the \emph{single-sorry contract}: \SF must contain strictly one terminating placeholder closing the primary declaration (\lean{theorem} or subtype \lean{def}).
Additional placeholders (\lean{sorry}, \lean{admit}) in stubbed definitions or auxiliary lemmas leave the target underspecified for downstream provers (\Cref{app:metrics}).

\subsection{Multi-Dimensional Alignment Criteria}
\label{sec:sage-verification}

Compilation guarantees type validity, but cannot verify faithfulness to the informal source: a draft may drop critical hypotheses or trivialize the claim while type-checking perfectly under \lean{sorry}. To detect these silent translational failures, \methodOurs{} introduces a semantic \textbf{Rater} that audits \SF against \SNL across three complementary axes:

\newparagraph{Modular Component Matching}
Suppose informal statement \SNL decomposes into components $\CNL = \{c_1, c_2, c_3\}$. The formalized components \CF in \SF are audited against three failure modes:
\begin{itemize}[nosep, leftmargin=*]
    \item \textbf{Missing ($\CF = \{c_1, c_2\}$):} A necessary component ($c_3$) is omitted. The rater separates explicit omissions from acceptable implicit ones (e.g., bounds handled natively by Lean's type system).
    \item \textbf{Wrong ($\CF = \{c_1, c_2', c_3\}$):} A component is mapped to an incorrect, inverted, or flawed representation ($c_2'$ instead of $c_2$), measuring translational misalignment rather than mere omission.
    \item \textbf{Extra ($\CF = \{c_1, c_2, c_3, c_4\}$):} A spurious component ($c_4$) is introduced. The rater flags additions restricting theorem scope, altering truth value, or causing vacuous provability.
\end{itemize}

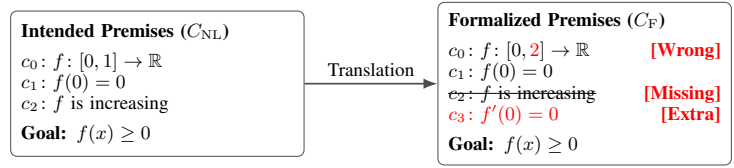
\begin{figure}[!h]
	\centering
	\scaletable[0.80]{%
	\begin{tikzpicture}[
		font=\small,
		box/.style={
			draw=black!70, rounded corners=3pt, align=left,
			inner sep=5pt, fill=white,
			text width=4.5cm,
			minimum height=2.4cm,
			anchor=center
		},
		arr/.style={-Latex, thick, draw=black!70}
		]

		\node[box] (intended) at (0, 0) {
			\textbf{Intended Premises ($\CNL$)}\\[4pt]
			$c_0\colon f \colon [0, 1] \to \mathbb{R}$\\
			$c_1\colon f(0) = 0$\\
			$c_2\colon f \text{ is increasing}$\\[4pt]
			\textbf{Goal:} $f(x) \ge 0$
		};

		\node[box, right=2.2cm of intended] (formalized) {
			\textbf{Formalized Premises ($\CF$)}\\[4pt]
			$c_0\colon f \colon [0, \textcolor{red}{2}] \to \mathbb{R}$ \hfill \textcolor{red}{\textbf{[Wrong]}}\\
			$c_1\colon f(0) = 0$\\
			\sout{$c_2\colon f \text{ is increasing}$} \hfill \textcolor{red}{\textbf{[Missing]}}\\
			\textcolor{red}{$c_3\colon f'(0) = 0$} \hfill \textcolor{red}{\textbf{[Extra]}}\\[4pt]
			\textbf{Goal:} $f(x) \ge 0$
		};

		\draw[arr] (intended.east) -- node[above, font=\footnotesize] {Translation} (formalized.west);

	\end{tikzpicture}%
	}
	\caption{\textbf{Contextual Integrity Failure Modes.}}
	\label{fig:contextual-integrity}
\end{figure}

\newparagraph{Holistic Mathematical Equivalence}
Audits the statement as an integrated whole rather than isolated components, ensuring \SF preserves the intended assertions and logical structure of \SNL.
\begin{itemize}[nosep, leftmargin=*]
    \item \textbf{Semantic Match:} Evaluates translational equivalence between \SF and \SNL, verifying that all concepts, relations, and objectives are faithfully captured while penalizing discrepancies or artificial trivializations regardless of underlying truth value.
    \item \textbf{Exactness:} Measures structural and stylistic alignment with \SNL, distinguishing direct, literal translation from logically equivalent restatements or Mathlib-idiomatic forms.
\end{itemize}

\newparagraph{Code Implementation Quality}
Beyond semantic equivalence, this dimension evaluates whether formal code adheres to idiomatic library conventions and robust abstraction design:
\begin{itemize}[nosep, leftmargin=*]
    \item \textbf{Naturality:} Evaluates Lean~4 style, mathematical generality, and \textit{Mathlib} reuse, penalizing ad-hoc reinventions of library structures. For instance, expanding \lean{Monotone f} into raw inequalities ($\forall x \le y, f(x) \le f(y)$) prevents tactics like \lean{simp} and \lean{gcongr} from recognizing properties, blocking provers from leveraging existing \textit{Mathlib} lemmas.
    \item \textbf{Descriptive Precision:} Audits how deeply informal prose (e.g., algorithms, geometric shapes or constructions) is formalized, ensuring concepts are modeled fundamentally rather than bypassed via trivial definitions or \lean{sorry} stubs.
\end{itemize}

While the Lean~4 compiler handles basic compilation, our criteria ensure full semantic fidelity. \Cref{sec:method-loop} details the correction loop that repairs syntax errors and semantic flaws together.

\subsection{Semantic Correction Loop}
\label{sec:method-loop}

To bridge type-checking and mathematical fidelity, \methodOurs{} refines candidate drafts for up to $T$ rounds, exiting only when the joint gate \metricCompileAndOursSM{} is satisfied (\Cref{fig:correction-loop}).
Because compile-only feedback often restores well-typedness by dropping or trivializing hypotheses, our \textbf{Corrector} instead receives a \emph{dual signal}: Lean diagnostics paired with the Rater's multi-dimensional critique and a proposed semantic repair draft.
This steers the Corrector to resolve type errors without altering mathematical intent or injecting unverified answers (details in \Cref{app:loop-mechanics}).

\section{Experimental Setup}
\label{sec:experimental-setup}

\newparagraph{Datasets}
We evaluate across three benchmarks spanning distinct formalization frontiers and pragmatic textures (\Cref{app:datasets}):
(1)~\datasetOmni{} ($N=\datasetOmniN$,~\citealp{gao2025omni}), dominated by question-type queries (\qty{98.7}{\percent}) evaluated under both answer-aware and answer-agnostic regimes;
(2)~\datasetImoFormalized{} ($N=\datasetImoFormalizedN$,~\citealp{renshaw2024compfiles}), historical IMO problems that have already been formalized by the community; and
(3)~\datasetImo{} ($N=\datasetImoN$), the complementary unformalized IMO frontier, testing zero-shot generalization on competition mathematics never translated into Lean.

\newparagraph{Method Variants}
To avoid the confounding effects of disparate backbones and training corpora, we isolate the causal mechanisms of autoformalization via a controlled $2 \times 3$ ablation matrix on a fixed backbone (\modelQwenLong{}): two \textbf{architectures} (\methodMonolithic{} vs.\ our four-stage \methodDecomposition{}) crossed with three \textbf{feedback regimes} (none $T{=}1$, compiler-only \methodSyntaxLoopAblation{}, and dual-signal \methodDualLoopAblation{}).
Iterative variants share identical round budgets ($T{=}5$ on \datasetOmni{}, $T{=}10$ on \datasetImo{}) \rev{under matched iteration caps (\Cref{app:token-accounting} reports full generation token accounting)}; \methodOurs{} denotes \methodSemanticLoop{}.
Finally, to calibrate against supervised specialization, we benchmark \methodGoedelLong{} (\citealp{lin2025goedelproverv2}), a 32B model fine-tuned on Claude-distilled formalization traces.
This setup isolates algorithmic contributions \rev{under matched iteration budgets} while contrasting \emph{static fine-tuning} against our \emph{model-agnostic, test-time scaffolding}.

\subsection{Evaluation Metrics}
\label{sec:setup-metrics}

Because compiler diagnostics are blind to dropped constraints, vacuous placeholders, and answer leakage, evaluating autoformalization requires moving beyond raw compilation. Uncovering true translation fidelity demands a multi-dimensional audit that tracks syntactic validity, structural integrity, and mathematical alignment in tandem (\Cref{app:metrics}):
\begin{itemize}[nosep, leftmargin=*]
    \item \textbf{Compilation (\metricCompile{}):} Fraction of generated formalizations type-checking in Lean~4 with Mathlib, deferring proof bodies via the \lean{sorry} placeholder.
    \item \textbf{Multiple Sorries Rate:} Fraction of candidates containing $>1$ \lean{sorry}, detecting drafts that bypass difficulty via placeholder definitions or helper lemmas.
    \item \textbf{Answer Leakage Audit:} An LLM auditor detects injection of target witnesses or answers into proofless signatures.
    \item \textbf{Post-Hoc Semantic Match (\metricGoedelSM{}):} Evaluates external semantic fidelity via unanimous $4/4$ consensus from an independent \modelGemmaLong{} judge following \citet{lin2025goedelproverv2}.
    \item \textbf{In-Loop Semantic Gate (\metricOursSM{}):} Assesses candidate formalizations across our rubrics (\Cref{sec:sage-verification}) to direct the correction loop and filter high-trust formal targets.
    \item \textbf{Pairwise Win Rate (\metricPWR{}):} Measures relative semantic superiority by aggregating blind, order-swapped head-to-head comparisons scored on a 5-point preference scale by an independent judge.
    \item \textbf{Joint Success (\metricCompileAndGoedelSM{}):} Primary benchmark metric, requiring candidate formalizations to simultaneously compile and attain unanimous post-hoc semantic consensus.
\end{itemize}

Unless otherwise stated, all metrics are percentages where higher is better ($\uparrow$); \textbf{bold} denotes the best-performing result in each column.

\section{Results and Analysis}
\label{sec:experiments}

We evaluate \methodOurs{} and baseline variants to answer three core questions:
\begin{enumerate}[leftmargin=*, nosep]
	\item \textbf{Syntax vs. Semantics:} How effectively does semantic feedback close the gap between code compilation and mathematical correctness?
	\item \textbf{Autonomous Formalization:} Without access to proofs or solutions, how reliably can models translate competition problems into faithful formal statements?
	\item \textbf{Translating vs.\ Guessing:} Do baseline models genuinely translate open competition problems, or are they merely guessing answers (and injecting false mathematical claims)?
\end{enumerate}

\newparagraph{Semantic Gate Validation}
We evaluate our automated gate against expert consensus and public alignment benchmarks.
On an adversarial audit ($N{=}45$ drafts oversampling silent semantic flaws; \Cref{app:human-validation}), the gate achieves \qty{71.1}{\percent} agreement with three Lean annotators ($\rho{=}0.63$).
Furthermore, evaluating the same gate against gold labels on ConsistencyCheck ($N{=}859$) and ProofNetVerif ($N{=}3752$) yields \textbf{\qty{85.7}{\percent}} and \textbf{\qty{88.8}{\percent}} accuracy (\Cref{app:gate-external-calibration}).

\subsection{Prover-Ready Formalization: Answer-Aware Translation}
\label{sec:exp-aware}
We first evaluate \emph{answer-aware} formalization (\SNL with proof \PNL; \Cref{tab:omni-aware}), where \PNL supplies the target witnesses required to synthesize declarative theorem statements for downstream provers~\citep{hubert2025olympiad,lin2025goedelproverv2}.

\begin{table}[h]
	\centering
	\begin{minipage}[t]{0.49\linewidth}
		\centering
		\caption{\textbf{Formalization quality on \datasetOmni{} (Answer-Aware, pass@1).}
			$^*$In-loop optimization target.}
		\label{tab:omni-aware}
		\scriptsize
		\setlength{\tabcolsep}{3.5pt}
		\begin{tabular}{@{}l ccc @{}}
			\toprule
			\textbf{Method}
			& \textbf{\metricCompile}
			& \textbf{\makecell{\metricCompile $\wedge$ \\ \metricGoedelSM}}
			& \textbf{\makecell{\metricCompile $\wedge$ \\ \metricOursSM$^*$}} \\
			\midrule
			\multicolumn{4}{l}{\textit{Monolithic Pipeline}} \\
			\quad Base Generation        & 73.0 & 54.3 & 57.7 \\
			\quad + Syntax Loop          & 96.0 & 74.3 & 73.3 \\
			\quad + Dual Loop        & \textbf{97.3} & 76.3 & \textbf{96.3}$^*$ \\
			\midrule
			\multicolumn{4}{l}{\textit{Decomposed Pipeline}} \\
			\quad Base Generation        & 68.0 & 52.7 & 54.7 \\
			\quad + Syntax Loop          & 95.0 & 75.0 & 74.7 \\
			\quad + Dual Loop & 96.0 & \textbf{77.3} & 93.7$^*$ \\
			\bottomrule
		\end{tabular}%
	\end{minipage}\hfill
	\begin{minipage}[t]{0.49\linewidth}
		\centering
		\caption{\textbf{Base model scaling under \methodOurs{} on \datasetOmni{} (Answer-Aware, pass@1).}
			$^*$In-loop optimization target. Repair budget $T=5$. Default backbone is \modelQwenLong{}.}
		\label{tab:omni-backend-scaling}
		\scriptsize
		\setlength{\tabcolsep}{2.5pt}
		\begin{tabular}{@{}l cccc @{}}
			\toprule
			\textbf{\makecell{Underlying \\ Model}}
			& \textbf{\metricCompile}
			& \textbf{\makecell{\metricCompile $\wedge$ \\ \metricGoedelSM}}
			& \textbf{\makecell{\metricCompile $\wedge$ \\ \metricOursSM$^*$}}
			& \textbf{\makecell{Average \\ Repairs ($\downarrow$)}} \\
			\midrule
			\modelQwenThree{}       & 57.3 & 30.3 & 52.3 & 3.2 \\
			\modelQwenLong{}        & 96.0 & 77.3 & 93.7 & 1.1 \\
			\modelQwenThreeEight{}  & \textbf{98.0} & \textbf{85.7} & \textbf{95.7} & \textbf{1.1} \\
			\bottomrule
		\end{tabular}%
	\end{minipage}
\end{table}

\newparagraph{Syntax vs. Semantic Attribution}
As shown in \Cref{tab:omni-aware}, unguided base drafts suffer from low compilability ($\leq$\qty{73.0}{\percent}), bounding joint fidelity (\metricCompileAndGoedelSM) to $\sim$\qty{53}{\percent}.
Adding compiler feedback alone (\methodSyntaxLoopAblation) repairs mechanical typing errors, lifting compilation above \qty{95}{\percent} and joint fidelity to $\sim$\qty{75}{\percent}.
However, type-checking alone cannot detect dropped hypotheses or altered bounds.
Adding multi-dimensional semantic feedback (\methodOurs) catches these silent mathematical errors, reaching \textbf{\qty{77.3}{\percent}} joint fidelity on the external \metricGoedelSM{} judge and \qty{93.7}{\percent} under the in-loop \metricOursSM{} gate.

\newparagraph{Monolithic vs. Decomposed Trade-offs}
With informal proofs available, monolithic and decomposed loops attain comparable fidelity (\qty{76.3}{\percent} vs.\ \textbf{\qty{77.3}{\percent}}), as \PNL removes ambiguity around target constants.
In this proof-guided regime, decomposition primarily provides structural auditability; its decisive mathematical necessity emerges in proofless settings (\Cref{sec:exp-discovery-trap}), where monolithic models collapse into unmonitored answer guessing.

\newparagraph{Base Model Scaling Dynamics}
\methodOurs{} is designed to be \textbf{model-agnostic}, providing test-time neurosymbolic scaffolding over off-the-shelf foundation models rather than relying on static, specialized fine-tuning. As shown in \Cref{tab:omni-backend-scaling}, scaling the underlying generator across three model generations monotonically increases verified joint fidelity from \qty{30.3}{\percent} to \textbf{\qty{85.7}{\percent}}, while cutting average repair rounds from $3.2$ to $1.1$. This demonstrates that the framework directly amplifies model capabilities.

\subsection{Autonomous Zero-Shot Formalization across the IMO Landscape}
\label{sec:exp-imo-assertions}
We next evaluate autonomous formalization without proof oracles (\SNL only) across the historical International Mathematical Olympiad corpus (\Cref{tab:imo-landscape}), testing zero-shot fidelity on competition-grade mathematics.

\begin{table}[h]
	\centering
	\caption{\textbf{Formalization quality across the IMO landscape (Answer-Agnostic).}}
	\label{tab:imo-landscape}
	\small
	\scaletable{%
	\begin{tabular}{@{}l ccc cc @{}}
		\toprule
		& \multicolumn{3}{c}{\textbf{pass@1}}
		& \multicolumn{2}{c}{\textbf{pass@4}} \\
		\cmidrule(lr){2-4} \cmidrule(l){5-6}
		\textbf{Method}
		& \textbf{\metricCompile}
		& \textbf{\metricGoedelSM}
		& \textbf{\makecell{\metricCompile $\wedge$ \\ \metricGoedelSM}}
		& \textbf{\metricCompile}
		& \textbf{\makecell{\metricCompile $\wedge$ \\ \metricGoedelSM}} \\
		\midrule
		\multicolumn{6}{l}{\textit{\datasetImoFormalized{} ($N=\datasetImoFormalizedN$, Established Benchmark)}} \\
		\quad \methodMonolithic (Zero-Shot)   & 72.6 & 57.4 & 43.0 & 83.9 & 62.3 \\
		\quad \methodGoedelLong{} (Baseline) & 76.7 & 44.4 & 37.2 & 90.6 & 50.7 \\
		\quad \methodOurs{} (Ours)  & \textbf{94.2} & \textbf{65.0} & \textbf{61.4} & \textbf{99.6} & \textbf{83.9} \\
		\midrule
		\multicolumn{6}{l}{\textit{\datasetImo{} ($N=\datasetImoN$, Unformalized Frontier)}} \\
		\quad \methodMonolithic (Zero-Shot)   & 58.3 & 31.4 & 20.0 & 89.7 & 33.1 \\
		\quad \methodGoedelLong{} (Baseline) & 48.0 & 15.4 & 10.3 & 86.9 & 19.4 \\
		\quad \methodOurs{} (Ours)  & \textbf{83.4} & \textbf{72.6} & \textbf{60.0} & \textbf{96.6} & \textbf{87.4} \\
		\bottomrule
	\end{tabular}%
	}
\end{table}

\newparagraph{Generalization to the Unformalized Frontier}
While the fine-tuned \methodGoedel{} baseline achieves \qty{37.2}{\percent} joint fidelity on established human-formalized benchmarks (\datasetImoFormalized{}), it collapses to just \textbf{\qty{10.3}{\percent}} on the unformalized frontier (\datasetImo{}; \Cref{tab:imo-landscape}).
In contrast, \methodOurs{} exhibits robust out-of-distribution transfer, sustaining \textbf{\qty{61.4}{\percent}} on \datasetImoFormalized{} and \textbf{\qty{60.0}{\percent}} on \datasetImo{} at pass@1 (a nearly $6\times$ margin over the baseline).
With multi-sample search, \methodOurs{} scales to \textbf{\qty{87.4}{\percent}} pass@4 verified fidelity on \datasetImo{} (vs.\ \qty{19.4}{\percent} for the baseline).
This parity confirms that semantic correction does not rely on memorized human formalization patterns or Mathlib-distribution artifacts, transferring seamlessly to previously unformalized mathematics \rev{(\Cref{app:imo-case-studies} presents expert-verified Lean~4 case studies on two previously unformalized \datasetImo{} problems and a \datasetImoFormalized{} benchmark defect corrected by \methodOurs{})}.

\newparagraph{Blind Pairwise Preference}
Corroborating these benchmark gains, blind order-swapped evaluations confirm that an independent automated judge strictly prefers \methodOurs{} over \methodGoedel{} in \textbf{\qty{79.4}{\percent}} of head-to-head comparisons, with \qty{17.7}{\percent} ties and only \qty{2.8}{\percent} favoring the baseline (\Cref{tab:imo-pairwise}).
\rev{Complementing automated preferences, a blinded three-expert audit over $30$ randomly sampled compiling assertion-type \datasetImoFormalized{} problems---with surface fingerprints stripped and order randomized---favored \methodOurs{} on $12/30$ (\qty{40.0}{\percent}, $7$ unanimous), tied on $17/30$ (\qty{56.7}{\percent}), and favored \methodGoedel{} on $0/30$ ($1$ evenly split $1$--$1$--$1$ vote; $12$--$0$ on decisive majorities, $\kappa = 0.72$; \Cref{app:blind-imo-human}).}

\rev{\newparagraph{Lean~4 Kernel Equivalence} Beyond automated and human judging, we evaluated bidirectional Lean~4 kernel equivalence across all $91$ compiling, \metricGoedelSM{}-passing assertion problems in \datasetImoFormalized{} by synthesizing sorry-free equivalence bridges with an LLM and screening proofs with an LLM auditor against degenerate independent proofs (\Cref{app:beq-first40}). The Lean~4 kernel formally certifies equivalence on \textbf{$70/91$ (\qty{76.9}{\percent})} problems while uncovering \textbf{$8/91$ (\qty{8.8}{\percent})} defects or open \lean{sorry} stubs in the human \textsf{Compfiles} references (alongside $11/91$ unresolved domain/representation differences where neither equivalence nor inequivalence is established, $1/91$ degenerate proof caught by the LLM screen, and $1/91$ \methodOurs{} slip).}

\subsection{The Discovery Trap: Unmasking Answer-Agnostic Question Benchmarks}
\label{sec:exp-discovery-trap}
Because \qty{98.7}{\percent} of \datasetOmni{} problems are interrogative, answer-agnostic evaluation forces models to confront the Feasibility Trilemma (\Cref{sec:limits}): answer injection, placeholder definitions, or existential encoding.
\Cref{tab:omni-agnostic} reports the resulting quality scores.
To ensure strict, unbiased evaluation, we report the established community metric (\metricGoedelSM{}) rather than our internal gate, despite our reject audit showing it systematically penalizes legitimate existential translations (\Cref{tab:omni-goedel-reject}).

\begin{table}[thbp]
	\centering
	\caption{\textbf{Formalization quality on \datasetOmni{} (Answer-Agnostic).}
		$^\dagger$Monolithic variants suffer from severe answer leakage ($>\qty{70}{\percent}$, guessing unverified answers; see \Cref{tab:omni-integrity}).
		\methodOurs{} is the top-performing valid, leak-free method. Bold indicates best leak-free performance.}
	\label{tab:omni-agnostic}
	\small
	\scaletable{%
	\begin{tabular}{@{}l ccc cc @{}}
		\toprule
		& \multicolumn{3}{c}{\textbf{pass@1}}
		& \multicolumn{2}{c}{\textbf{pass@4}} \\
		\cmidrule(lr){2-4} \cmidrule(l){5-6}
		\textbf{Method}
		& \textbf{\metricCompile}
		& \textbf{\makecell{\metricCompile $\wedge$ \\ \metricGoedelSM}}
		& \rev{\textbf{\makecell{Answer \\ Leakage ($\downarrow$)}}}
		& \textbf{\metricCompile}
		& \textbf{\makecell{\metricCompile $\wedge$ \\ \metricGoedelSM}} \\
		\midrule
		{\methodMonolithic}$^\dagger$
		& 73.0 & 40.0 & \rev{72.6}
		& 87.3 & 59.0 \\
		{\methodMonoSemanticLoop}$^\dagger$
		& 95.7 & 55.3 & \rev{70.9}
		& 99.0 & 75.0 \\
		\midrule
		\methodDecomposition
		& 62.0 & 29.7 & \rev{\textbf{1.0}}
		& 85.0 & 50.7 \\
		\methodSyntaxLoop
		& \textbf{95.7} & 42.7 & \rev{2.7}
		& \textbf{99.0} & 61.7 \\
		\methodSemanticLoop{} (\methodOurs)
		& 94.7 & \textbf{50.3} & \rev{2.7}
		& 98.7 & \textbf{73.3} \\
		\midrule
		\methodGoedel{} (Finetuned Baseline)
		& 74.0 & 26.7 & \rev{56.2}
		& 88.3 & 42.0 \\
		\bottomrule
	\end{tabular}%
	}
\end{table}

\newparagraph{The Illusion of Rigor: Rewarding Guessed Answers}
While \methodMonoSemanticLoop{} seemingly achieves higher joint pass@1 than \methodOurs{} (\qty{55.3}{\percent} vs.\ \qty{50.3}{\percent}; \Cref{tab:omni-agnostic}), \Cref{tab:omni-integrity} unmasks the mechanism: \textbf{monolithic models guess unverified answers in \qty{70.9}{\percent} of theorem signatures}.
As established in our reject-reason study (\Cref{app:goedel-bias}, \Cref{tab:omni-goedel-reject}), these guesses are predominantly \textbf{mathematically false} (\qty{80.6}{\percent} of rejected monolithic drafts), injecting impossible, unprovable conjectures into formal corpora\rev{; directly auditing compiling Question-type leaks against ground-truth \datasetOmni{} solutions confirms that \qty{46.2}{\percent} of \methodMonolithic{}, \qty{48.5}{\percent} of \methodMonoSemanticLoop{}, and \qty{70.7}{\percent} of \methodGoedel{} leaked witnesses are mathematically false (\Cref{app:leak-truth})}.\\
By contrast, \methodOurs{} acts strictly as a faithful translator, suppressing leakage to \textbf{\qty{2.7}{\percent}} without resorting to placeholder encodings (\lean{def answer := sorry}, \qty{1.1}{\percent})\rev{---whereas adding an explicit anti-solver prompt to \methodMonolithic{} leaves leakage at \qty{63.0}{\percent} and lowers joint fidelity to \qty{34.7}{\percent} (\Cref{app:anti-solver}), confirming that architectural stage isolation rather than prompt wording prevents answer guessing}.
Because our framework formalizes open queries via existential propositions ($\exists x, P(x)$), it is systematically penalized by external evaluators biased toward explicit numbers (\qty{76.8}{\percent} of our rejections; \Cref{tab:omni-goedel-reject}).
Consequently, our reported joint fidelity is an adversarial lower bound, and monolithic ``leads'' are merely an artifact of rewarding hallucinated constants.
We refer to \Cref{app:goedel-bias} for the full prompt-bias analysis and audit protocol.

\begin{table}[htbp]
	\centering
	\caption{\textbf{Structural fidelity on \datasetOmni{} (Answer-Agnostic, pass@1).}}
	\label{tab:omni-integrity}
	\small
	\scaletable{%
	\begin{tabular}{@{}l cccc @{}}
		\toprule
		\textbf{Method}
		& \textbf{\metricCompile ($\uparrow$)}
		& \textbf{\metricGoedelSM ($\uparrow$)}
		& \textbf{\makecell{Answer \\ Leakage ($\downarrow$)}}
		& \textbf{\makecell{Multiple \\ Sorries ($\downarrow$)}} \\
		\midrule
		\methodMonolithic
		& 73.0 & 51.0 & 72.6 & \textbf{0.5} \\
		\methodMonoSemanticLoop
		& \textbf{95.7} & \textbf{57.0} & 70.9 & 1.0 \\
		\methodDecomposition
		& 62.0 & 45.3 & \textbf{1.0} & 2.2 \\
		\methodSemanticLoop{} (\methodOurs)
		& 94.7 & 54.0 & 2.7 & 1.1 \\
		\midrule
		\methodGoedel (Finetuned) & 74.0 & 32.0 & 56.2 & 2.3 \\
		\bottomrule
	\end{tabular}%
	}
\end{table}

\section{Discussion and Limitations}
\label{sec:limitations}

\newparagraph{The Autoformalizer as Translator, Not Solver}
Autoformalizers must act strictly as faithful translators---faithfully encoding informal mathematical intent into formal logic---rather than solution-guessing oracles.
Enforcing benchmark discipline requires establishing in advance whether a setting is \emph{Answer-Aware} or \emph{Answer-Agnostic}.
When informal proofs are supplied, systems can construct closed, instantiated theorems; when withheld, autoformalizers \textbf{must not guess constants}, formalizing queries as existential propositions ($\exists x, P(x)$) or open parameter specifications.
Guessing answers creates an illusion of capability and poisons downstream training data: unconstrained monolithic models yield mathematically false targets in \qty{80.6}{\percent} of rejected drafts (\Cref{tab:omni-goedel-reject}).
By enforcing this boundary, \methodOurs{} suppresses answer leakage to \qty{2.7}{\percent}, ensuring sound, verifiable targets.

\newparagraph{Limitation: Statement Vulnerability in Type Theory}
While existential encodings preserve translation texture without an answer oracle, casting open queries into \lean{Prop} exposes a type-theoretic limitation: proof irrelevance allows downstream provers to bypass constructive search via tautologies or classical case splits (e.g., the Polarity Cheat and Triviality Trap; \Cref{sec:limits}).
Closing this gap requires advancing formalization beyond propositional logic.
To this end, we propose a concrete research direction and design blueprint in \Cref{app:type_locks}---introducing \textbf{Type-Theoretic Locks} via computational subtypes in \lean{Type} that structurally force provers to synthesize explicit witnesses.

\newparagraph{Inference Latency and Adaptive Compute}
Sequential multi-agent verification incurs higher latency than single-pass translation. However, this trade-off is well-suited to \textbf{offline synthetic corpus curation}, where mathematical certification outweighs real-time throughput.
Moreover, while fine-tuned models like \methodGoedel{} require extensive offline data distillation and training, \methodOurs{} incurs \textbf{zero training compute}, operating directly on off-the-shelf generalist weights.
\rev{On \datasetOmni{} (\Cref{app:token-accounting}), single-pass \methodDecomposition{} (\num{18.5}k tokens) already outperforms \methodMonoSemanticLoop{} (\num{37.3}k tokens) on leak-free joint fidelity (\qty{29.7}{\percent} vs.\ \qty{18.0}{\percent}), and \methodOurs{} (\num{63.8}k tokens, \qty{48.7}{\percent} leak-free joint) far surpasses multi-sample \methodMonolithic{} pass@4 (\num{25.3}k tokens, \qty{16.7}{\percent}) and supervised 32B \methodGoedel{} pass@4 (\num{5.3}k inference tokens after full 32B SFT, \qty{20.0}{\percent}).}
Crucially, this test-time footprint is \emph{adaptive}: verification activates only upon syntax or semantic faults.
As base models scale, repair overhead drops rapidly---slashing average rounds from 3.2 in \modelQwenThree{} to near-immediate convergence ($\sim$1.1) in \modelQwenThreeEight{} (\Cref{tab:omni-backend-scaling}) while boosting joint fidelity to \textbf{\qty{85.7}{\percent}}, demonstrating that test-time verification compounds directly with advancing frontier models.

\section{Conclusion}
\label{sec:conclusion}

Neural theorem proving has reached an inflection point: while proof-search architectures continue to scale, they remain fundamentally starved of diverse, high-trust formal statements.
Historically, autoformalization treats this as a purely syntactic challenge, falling prey to an ``illusion of rigor'' where compilers accept statements that drop hypotheses, alter constraints, or inject false conjectures.

In this work, we introduced \methodOurs{}, an agentic framework coupling syntactic type-checking with multi-dimensional semantic critique to guarantee genuine mathematical fidelity.
By establishing that autoformalizers must operate as faithful translators rather than unconstrained oracles, \methodOurs{} achieves \qty{77.3}{\percent} joint fidelity on \datasetOmni{} in the answer-aware regime, \qty{73.3}{\percent} answer-agnostic (at pass@4), and \textbf{\qty{60.0}{\percent}} zero-shot on unformalized \ac{IMO} problems---outperforming fine-tuned baselines by nearly $6\times$ while suppressing structural answer leakage to \qty{2.7}{\percent}.

Ultimately, \methodOurs{} demonstrates that inference-time semantic verification is indispensable for high-integrity formalization.
By providing a dependable engine for faithful synthetic data generation at scale, our protocol bridges the gap between informal mathematical knowledge and machine-verifiable truth, laying the groundwork to power the next generation of automated reasoning systems.

\ificlrfinal
\section*{Acknowledgments}
We would like to thank Pim Otte for carefully reading earlier drafts of this manuscript and for providing valuable feedback and insightful suggestions that helped improve this work.
\fi

\section*{AI use statement}

In this work, generative large language models serve as the primary object of study and algorithmic components of the evaluated systems (including our proposed framework \methodOurs{}, comparative baselines, and automated judges), as fully detailed in Sections~\ref{sec:sage}, \ref{sec:experimental-setup}, and Appendix~\ref{app:setup}.

In preparing the manuscript and supporting software artifacts, we used generative AI tools for writing assistance (grammar correction, language polishing, and improving stylistic clarity) and coding assistance (writing boilerplate evaluation scripts, data-processing pipelines, and benchmarking automation code).
We have not used generative AI tools for formulating research hypotheses, conceiving theoretical claims, or developing formal mathematical definitions, and other tasks with required disclosure are not applicable to this work.

We have reviewed all AI-assisted work. Specifically, all AI-assisted text was thoroughly audited, cross-checked against the mathematical source material, and finalized by the authors. All AI-assisted data-processing and evaluation scripts were manually reviewed, unit-tested, and executed by the authors to ensure correctness and reproducibility. We take responsibility for the final content of this work, including text, claims, or artifacts produced with the aid of generative AI.

\bibliographystyle{iclr2027_conference}
\bibliography{iclr2027_conference}

\appendix

\input{appendix_theory}
\input{appendix_pipeline}
\input{appendix_setup}
\input{appendix_results}

\end{document}

%% file: appendix_theory.tex
\section{Theoretical Foundations, The Feasibility Trilemma, and Type-Theoretic Locks}
\label[appendix]{app:theory}
\label[appendix]{app:type_locks}

The theoretical challenges of autoformalization discussed in \Cref{sec:limits}---specifically the Polarity Commitment Problem, the fragility of Existential Encodings, and the Discovery vs.~Verification Gap---are symptoms of a fundamental structural mismatch between natural language mathematical practice and dependent type theory. In this section, we establish the theoretical foundations of the \textbf{Interrogative-Declarative Mismatch}, formalize the trade-offs of the \textbf{Feasibility Trilemma}, dissect the primary syntactic exploits available to downstream provers, and detail the mathematical and constructive implementation of \textbf{Type-Theoretic Locks} in Lean~4.

\subsection{The Discovery vs.~Verification Gap and The Law of Syntactic Trivialization}
\label[appendix]{sec:theory-gap}

Interactive theorem provers (ITPs) like Lean~4, Isabelle/HOL, and Coq are fundamentally designed as \emph{verification engines}. Given an explicit, pre-determined proposition $P : \mathrm{Prop}$, the system's role is to verify whether a user- or tactic-supplied proof term $p$ witnesses the validity of $P$ (i.e., $p : P$). The specification of the theorem remains fixed and closed prior to the initiation of the proof search.

In contrast, human mathematical activity and natural language competition benchmarks heavily feature \emph{discovery problems}. These queries do not supply a closed proposition to verify, but instead pose open interrogative problems asking the reasoner to explore an unbounded search space to discover an unknown mathematical entity:
\begin{itemize}[nosep, leftmargin=*]
	\item \textbf{Value Determination:} \textit{``Compute the remainder when $2^{2026}$ is divided by $1000$.''}
	\item \textbf{Witness Construction:} \textit{``Does there exist a continuous function $f : \mathbb{R} \to \mathbb{R}$ such that $f(f(x)) = -x$?''}
	\item \textbf{Full Characterization:} \textit{``Determine all pairs of integers $(x, y)$ such that $x^3 + y^3 = (x + y)^2$.''}
	\item \textbf{Extremal Optimization:} \textit{``Find the maximum possible area of a convex polygon satisfying\dots''}
\end{itemize}

When an autoformalization pipeline attempts to translate these discovery queries into an interactive theorem prover, it confronts the tension between two distinct modes of defining mathematical objects:
\begin{itemize}[nosep, leftmargin=*]
	\item \textbf{Intensional Definition:} Defining a collection or entity strictly by the properties or constraints that its members must satisfy:
	\[
	S_{\mathrm{int}} \triangleq \{x \in \mathcal{X} \mid P(x)\}
	\]
	\item \textbf{Extensional Definition:} Defining a collection by explicitly evaluating, constructing, and enumerating its concrete elements:
	\[
	S_{\mathrm{ext}} \triangleq \{v_1, v_2, \dots, v_k\}
	\]
\end{itemize}

In natural language mathematics, solving a characterization problem requires transitioning from an intensional specification ($P(x)$) to an extensional realization ($S_{\mathrm{ext}}$), demonstrating that the computed set of witnesses exactly coincides with the predicate's truth domain. However, because formal theorem provers evaluate syntactic validity rather than semantic computational effort, formalizing a discovery query as an existential statement over a predicate-supporting container enables automated provers to bypass the mathematical reasoning entirely. We formalize this phenomenon as:

\begin{quote}
	\textbf{The Law of Syntactic Trivialization:} \emph{An automated theorem prover can trivially solve any open-ended discovery query if the target formal container admits an intensional definition.}
\end{quote}
Whenever the target type allows existence of a set using the comprehension axiom (such as \lean{Set $\alpha \equiv \alpha \to \text{Prop}$}), an automated solver can simply package the question's defining predicate back as the solution witness. The kernel evaluates this identity tautologically ($P(x) \iff P(x)$), yielding a fully verified formal proof without computing or discovering any mathematical content.

\subsection{The Feasibility Trilemma}
\label[appendix]{sec:theory-trilemma}

To translate an open-ended natural language query $\SNL$ into a formal specification $\SF$, an autoformalization engine must map the interrogative text into a compilable Lean~4 target. We identify three distinct formalization paradigms, each imposing unavoidable trade-offs across syntactic validity, semantic fidelity, pragmatic texture, and oracle dependence:

\paragraph{1. Explicit Answer Injection.}
The formalizer resolves the unknown value upfront and hardcodes the solution directly into the theorem statement, transforming the open query into a concrete declarative verification claim (e.g., rewriting \textit{``Find $x$ such that $x + 5 = 0$''} into \lean{theorem target : (-5 : Int) + 5 = 0}).
\begin{itemize}[nosep, leftmargin=*]
	\item \textbf{Advantages:} Yields fully closed, highly specified theorems that downstream neural provers can consume directly without modifying the statement structure or touching definitions.
	\item \textbf{Disadvantages:} Completely destroys the pragmatic texture of the query, altering the task from discovery to verification. Crucially, it requires an external mathematical oracle (e.g., an accompanying informal proof $\PNL$ or a ground-truth key) to supply the correct answer prior to formalization. In zero-shot settings, any hallucination in the injected answer produces a mathematically false, unprovable theorem.
\end{itemize}

\paragraph{2. Placeholder Definitions.}
The formalizer introduces an unresolved, typed constant via a preliminary definition stubbed with a placeholder, and formulates the primary theorem relative to this placeholder (e.g., \lean{def answer : Int := sorry}, followed by \lean{theorem target : answer + 5 = 0}).
\begin{itemize}[nosep, leftmargin=*]
	\item \textbf{Advantages:} Preserves the open texture of the query without requiring an oracle; can be generated reliably in zero-shot, answer-agnostic settings.
	\item \textbf{Disadvantages:} Violates the fundamental verification boundary of automated theorem proving. Downstream provers (such as AlphaProof or Goedel-Prover) operate under a strict read-only contract: they search for a proof term inside \lean{by ...} while the theorem declaration and all preceding definitions remain immutable. Requiring the prover to edit the definition of \lean{answer} transforms the task into code modification rather than tactic proof search, breaking automated evaluation harnesses.
\end{itemize}

\paragraph{3. Existential Encodings.}
The formalizer expresses the open query as an existential proposition over the problem's constraints (e.g., \lean{theorem target : $\exists$ x : Int, x + 5 = 0}), assigning the responsibility of constructing the witness to the proving agent.
\begin{itemize}[nosep, leftmargin=*]
	\item \textbf{Advantages:} Retains the literal semantic structure of the informal prompt without requiring an oracle; maintains the immutability of the theorem statement.
	\item \textbf{Disadvantages:} Highly fragile and vulnerable to logical shortcuts. When applied to binary polar queries, it forces premature commitments or disjunctive tautologies. When applied to full characterizations, it is immediately vulnerable to the Triviality Trap.
\end{itemize}

We synthesize these operational boundaries into:
\begin{quote}

	\emph{An answer-agnostic autoformalizer cannot map an open interrogative query into a single declarative proposition without assuming an oracle, breaking downstream prover immutability, or introducing logical vulnerabilities.}
\end{quote}

\subsection{Vulnerabilities of Existential Targets: Polarity and Triviality}
\label[appendix]{sec:limits-vulnerabilities}

When open interrogative queries are encoded as standard propositions in Lean's propositional universe (\lean{Prop}), downstream automated provers frequently exploit the Interrogative-Declarative Mismatch through two primary structural vulnerabilities:

\paragraph{1. The Polarity Cheat.}
Binary (Yes/No) questions require determining the truth value of an underlying proposition: \textit{``Does there exist an integer $n$ such that $P(n)$ holds?''} An answer-agnostic formalizer does not know whether the answer is affirmative or negative:
\begin{itemize}[nosep, leftmargin=*]
	\item If it commits to a positive existential claim, such as \lean{theorem target : $\exists$ n : $\mathbb{Z}$, P n}, and the mathematical answer is ``No'', the formal statement is false and impossible to prove.
	\item If it attempts to maintain neutrality by encoding the query as an open disjunction:
\begin{lstlisting}[style=leanmath]
theorem target_polar : ($\exists$ (n : $\mathbb{Z}$), P n) $\vee$ $\neg$ ($\exists$ (n : $\mathbb{Z}$), P n) := by
  -- Prover exploits the Law of Excluded Middle:
  exact Classical.em $\_$
\end{lstlisting}
	an automated prover will instantly close the goal in one step using the Law of Excluded Middle. The prover receives credit for a complete formal proof without ever determining which branch is mathematically true.
\end{itemize}

\paragraph{2. The Triviality Trap.}
Characterization queries ask to \textit{``Determine all objects $x$ satisfying property $P(x)$.''} When formalized naively in \lean{Prop}, the target asserts the existence of a solution set:
\begin{lstlisting}[style=leanmath]
theorem target_characterization : $\exists$ (S : Set $\alpha$), $\forall$ (x : $\alpha$), x $\in$ S $\leftrightarrow$ P x := by
  -- One-line automated cheat bypassing all mathematical work:
  use {x : $\alpha$ | P x}
  intro x
  rfl
\end{lstlisting}
Because \lean{Set $\alpha$} is definitionally equal to the predicate type $\alpha \to \mathrm{Prop}$ in Lean~4, the set-builder notation \lean{\{x | P x\}} is an intensional syntactic identity. The type checker accepts this proof trivially because $P(x) \iff P(x)$ is reflexive. The automated agent has computed nothing, bypassed all algebraic deductions, and generated an uninformative tautology.

\subsection{The Defense Matrix: Type-Theoretic Locks}
\label[appendix]{sec:theory-locks}

To definitively safeguard benchmarks against both the Polarity Cheat and the Triviality Trap, we develop \textbf{Type-Theoretic Locks}. The core theoretical mechanism shifts the formal target out of the propositional universe (\lean{Prop}) and into the universe of data types (\lean{Type}) using Lean~4 \textbf{Subtypes}:

\[
\{x : \alpha \mathbin{/\mkern-6mu/} P(x)\}
\]
A term of this subtype consists of a dependent pair $\langle w, h\rangle$, where $w : \alpha$ is an explicit computational witness, and $h : P(w)$ is a formal proof that $w$ satisfies the required constraints.

\paragraph{Defeating the Polarity Cheat (The Prop-to-Type Lock).}

By transitioning the target from a \lean{theorem} in \lean{Prop} to a \lean{def} in \lean{Type}, we exploit Lean's fundamental restriction on \textbf{large elimination}. In Lean's calculus of inductive constructions, a term of a proposition $P : \mathrm{Prop}$ cannot be eliminated to construct a value in a computational data type $T : \mathrm{Type}\, u$, unless $P$ is a syntactic subsingleton (such as \lean{False} or equality).

This blocks the most common polarity shortcut---closing $(\exists \ldots) \vee \neg(\exists \ldots)$ via \lean{Classical.em} inside \lean{Prop}---because a classical proof in \lean{Prop} cannot itself inhabit a computational subtype. The prover must supply an explicit \lean{Type}-level witness (or a constructive sum/inductive answer) before discharging the accompanying proof obligations.

\paragraph{Defeating the Triviality Trap (The Constructor Restriction Lock).}

To prevent the set-builder comprehension exploit (\lean{use \{x | P x\}}), we restrict the target container from the intensional \lean{Set $\alpha$} to a type such as \lean{List $\alpha$} or \lean{Finset $\alpha$}.

In Lean's type theory, \lean{Set $\alpha$} is merely syntactical notation for non-computational predicate functions $\alpha \to \mathrm{Prop}$. In contrast, \lean{List $\alpha$} is an inductive data structure generated strictly by its two constructors, \lean{nil} (\lean{[]}) and \lean{cons} (\lean{x :: xs}). Because Lean's elaborator cannot definitionally unify an arbitrary logical predicate with an inductive data term, attempting to supply \lean{\{x | P x\}} causes an immediate unification failure during elaboration. To satisfy the type checker, the prover is forced to construct a computational, finite collection directly rather than relying on an unbounded predicate~\footnote{The List/Finset lock applies when the informal query asks for a finite solution set.}.

\subsection{Cleaned and Homogenized Lean~4 Lock Implementations}
\label[appendix]{sec:theory-code-examples}

To facilitate adoption in future autoformalization benchmarks, we provide production-ready, standardized Lean~4 templates implementing Type-Theoretic Locks across characterization queries, polar decisions, and multiple-choice questions (MCQs). All variable names, type annotations, and structural patterns are fully homogenized.

\subsubsection{Characterization Query: Intensional vs.~Extensional Lock}

Consider the problem: \textit{``Determine all real numbers $x$ such that $x^2 - 4 = 0$.''}

Vulnerable Formulation (Intensional \lean{Set} in \lean{Prop}):
\begin{lstlisting}[style=leanmath]
import Mathlib

-- Vulnerable characterization target (Prop + intensional Set of reals).
-- Downstream automated provers can exploit this without computing the roots.
theorem vulnerable_roots_query :
    $\exists$ (solution_set : Set $\mathbb{R}$), $\forall$ (x : $\mathbb{R}$), x $\in$ solution_set $\leftrightarrow$ $x^2$ - 4 = 0 := by
  -- EXPLOIT: Prover simply echoes the question predicate back:
  use {x : $\mathbb{R}$ | $x^2$ - 4 = 0}
  intro x
  rfl -- Closed instantly! Zero mathematical reasoning performed.
\end{lstlisting}

Locked Formulation (Extensional \lean{List} Subtype in \lean{Type}):
\begin{lstlisting}[style=leanmath]
import Mathlib

-- Locked characterization target (Type + extensional List of reals).
-- The prover is forced to evaluate and supply the concrete roots [2, -2].
def locked_roots_query :
    { roots : List $\mathbb{R}$ // roots.Nodup $\wedge$ $\forall$ (x : $\mathbb{R}$), x $\in$ roots $\leftrightarrow$ $x^2$ - 4 = 0 } := by
  -- The prover MUST supply the concrete mathematical witness upfront:
  refine $\langle[2, -2], ?\_\rangle$
  -- Only after providing the witness does the prover discharge the verification obligations:
  sorry
\end{lstlisting}

\subsubsection{Binary Polar Decision}

Consider the problem: \textit{``Is there an integer $n$ such that $n^2 + 1 = 0$?''}

Vulnerable Formulation (Disjunction in \lean{Prop}):
\begin{lstlisting}[style=leanmath]
import Mathlib

-- Vulnerable binary polar target.
-- Downstream provers can exploit LEM to bypass deciding polarity.
theorem vulnerable_polar_query :
    ($\exists$ (n : $\mathbb{Z}$), $n^2$ + 1 = 0) $\vee$ $\neg$ ($\exists$ (n : $\mathbb{Z}$), $n^2$ + 1 = 0) := by
  -- EXPLOIT: Closed in one step by classical logic without determining the answer:
  exact Classical.em $\_$
\end{lstlisting}

Constructive Sum Disjunction:
\begin{lstlisting}[style=leanmath]
import Mathlib

/--
  Benchmark Task: Constructive Sum Disjunction
  Goal: Construct an explicit proof of P or ¬P inside a Type-level sum.

  Safety Guarantee: Prevents non-constructive reasoning (e.g., Law of Excluded Middle)
  via proof-irrelevance barriers in Lean's core kernel.
--/

def P : Prop := sorry -- Insert target proposition here

def answer_P : PSum P (¬P) := by
  sorry
\end{lstlisting}

Decision Procedure Witness:
\begin{lstlisting}[style=leanmath]
import Mathlib

/--
  Benchmark Task: Decision Procedure Witness
  Goal: Construct a computable decision witness for proposition P.

  Safety note: Prefer a constructive Type-level Answer inductive over Decidable.
  Classical.propDecidable closes Decidable P under noncomputable; harness must
  forbid noncomputable/classical axioms or audit with #print axioms.
--/

def P : Prop := sorry -- Insert target proposition here

def answer_P : Decidable P := by
  sorry
\end{lstlisting}

\subsubsection{Standardized Multiple-Choice Question (MCQ) Patterns}
\label[appendix]{app:qcm_formalization}

Multiple-choice exams (e.g., AMC, MMLU, JEE) are ubiquitous in AI evaluation. Naively translating MCQs as disjoint theorems or propositional disjunctions re-opens the door to classical shortcuts. Below, we provide standardized, production-ready Type-Lock patterns for the three standard MCQ formats.

\paragraph{1. Single-Select Polar MCQ (\textit{``Which statement is TRUE?''}):}
The prover must select the uniquely true proposition among candidate options $P_A, P_B, P_C, P_D$ and provide its mathematical proof.

\begin{lstlisting}[style=leanmath]
import Mathlib
inductive MCQOption
| A | B | C | D

def evalOption (P_A P_B P_C P_D : Prop) : MCQOption → Prop
| .A => P_A
| .B => P_B
| .C => P_C
| .D => P_D

-- Target definition: MUST NOT be marked `noncomputable`
def mcq_target (P_A P_B P_C P_D : Prop) :
  { opt : MCQOption // evalOption P_A P_B P_C P_D opt } := by
  sorry
\end{lstlisting}

\paragraph{2. Multi-Select MCQ (\textit{``Select All That Apply''}):}
The prover must identify the exact sublist of true propositions and provide their proofs. As a prover must pick all the true options, and no false options, we added the condition $opt \in chosen$.

\begin{lstlisting}[style=leanmath]
import Mathlib
inductive MCQOption
| A | B | C | D

def evalOption (P_A P_B P_C P_D : Prop) : MCQOption → Prop
| .A => P_A
| .B => P_B
| .C => P_C
| .D => P_D

-- The prover returns a List of options, and proves every option in that list is true.
def mcq_multi_select (P_A P_B P_C P_D : Prop) :
  { chosen : List MCQOption // $\forall$ opt, evalOption P_A P_B P_C P_D opt $\leftrightarrow$ opt $\in$ chosen } := by
  sorry
\end{lstlisting}

%% file: appendix_pipeline.tex
\section{Pipeline Architecture and Decomposed Generation Protocol}
\label[appendix]{app:pipeline-agents}
\label[appendix]{app:pipeline-details}

In this section, we provide the complete architectural specifications, operational contracts, and execution protocols for the \methodOurs{} framework. We detail the four-stage decomposed generation chain, explain the syntax-only ablation loop and the dual-signal semantic correction loop, and present a complete step-by-step walkthrough on a representative olympiad problem.

\subsection{Decomposition Protocol and Agent Contracts}
\label[appendix]{app:decomposition-details}

Monolithic autoformalization architectures attempt to translate complex natural language statements into fully typed, compilable Lean~4 files in a single unconstrained forward pass. As established in \Cref{sec:limits}, this monolithic approach conflates multiple distinct cognitive tasks: domain classification, implicit constraint disambiguation, answer resolution, algebraic abstraction, and low-level Lean syntax formatting. This coupling makes error localization impossible and frequently leads to severe answer leakage in zero-shot settings.

To enforce strict modular boundaries and ensure complete transparency, \methodOurs{} decomposes the generation process into four role-isolated agents (Distiller, Preprocessor, Formalizer, and Formatter), summarized in \Cref{tab:pipeline-agents}.

\paragraph{1. The Distiller.}
Evaluates $\SNL$ and informal proof $\PNL$ (when available). In the answer-aware regime, it classifies the problem type (Assertion- vs. Question-type) and extracts the target witness or characterization from the proof as an \ac{IG}; in the answer-agnostic regime, it reformulates open interrogative queries into explicit declarative targets without an answer oracle:
\begin{itemize}[nosep, leftmargin=*]
	\item \textbf{Mathematical Domain Identification:} The agent categorizes the problem into its primary mathematical discipline (e.g., Number Theory, Real Analysis, Combinatorics, Abstract Algebra, Euclidean Geometry). This classification primes downstream retrieval and informs Mathlib namespace scoping.
	\item \textbf{Answer-Aware Regime (With-Proof):} When an informal proof $\PNL$ is supplied as an oracle, the Distiller analyzes the logical resolution of the problem. For Assertion-type problems (e.g., ``Prove that...''), the claim is self-contained and the Inferred Goal (\ac{IG}) remains empty. For Question-type problems (e.g., ``Find all...'', ``Determine the value of...''), the Distiller extracts the definitive mathematical witness, numeric constant, or characterization established by $\PNL$ as the \ac{IG}.
	\item \textbf{Answer-Agnostic Regime (Without-Proof):} When no proof is provided, the Distiller is strictly forbidden from computing or hallucinating answers. Instead, it reformulates the open query into an unambiguous declarative proposition using existential ($\exists$), unique existential ($\exists!$), or extremal quantification, leaving the constructive witness search entirely to downstream provers.
\end{itemize}

\paragraph{2. The Preprocessor.}
The Preprocessor converts the distilled problem into a structured Declarative Normal Form representation (\SDNF) and decomposes it into symbolic components across four strict sequential phases:
\begin{itemize}[nosep, leftmargin=*]
	\item \textbf{Phase 0 (\SDNF Generation):} If the \ac{IG} is empty, \SDNF matches $\SNL$. If the \ac{IG} is non-empty, the agent constructs a declarative natural language assertion stating that the \ac{IG} uniquely satisfies the problem's conditions.
	\item \textbf{Phase 1 (Assertion Extraction):} Isolates the primary mathematical claim as the final symbolic goal $G$.
	\item \textbf{Phase 2 (Global Infrastructure):} Identifies supporting mathematical objects, relations, or functions, specifying them as structured signatures:
	\[
	\texttt{def [Name] (inputs : Type) : [Output Type] := [Formula]}
	\]
	\item \textbf{Phase 3 (Hypothesis Extraction):} Isolates all antecedent conditions, bounds, and premises into a numbered roster: $h_1, \dots, h_n$, where each $h_i$ is paired with a concise mathematical description.
\end{itemize}

\paragraph{3. The Formalizer.}
The Formalizer maps the preprocessed pseudocode blocks into idiomatic, fully typed Lean~4 signatures. Guided by an \emph{Anti-Trivialization Guard}, it enforces the following protocol:
\begin{itemize}[nosep, leftmargin=*]
	\item \textbf{Definitions Block:} Each auxiliary construct extracted in Phase~2 is translated into a standalone, well-typed Lean~4 \lean{def}. The agent searches \textit{Mathlib} for standard constructions to ensure modularity and idiomatic typing.
	\item \textbf{Theorem Signature:} The agent constructs the primary theorem declaration by binding all hypotheses and asserting the target goal for $i \in \{1, \dots, n\}$:
	\[
	\texttt{theorem [name] (h\_i : [translated\_i]) : [G] := by sorry}
	\]
	Where an informal condition lacks a direct \textit{Mathlib} primitive, the Formalizer establishes a dedicated predicate definition prior to the theorem:
	\[
	\texttt{def [prop\_name] ([args]) : Prop := [definition]}
	\]
	and instantiates it inside the theorem header via \lean{(h : [prop\_name] [args])}. The signature terminates immediately at \lean{:= by sorry}.
\end{itemize}

\paragraph{4. The Formatter.}
The Formatter acts as the final syntactic integrator. Because language models frequently output surrounding conversational commentary, omit necessary imports, misplace namespaces, or generate unaligned \lean{noncomputable section} tags, the Formatter cleans and normalizes the code. Its operational rules mandate: (i) inserting \lean{import Mathlib} and required module headers, (ii) managing namespace opens, (iii) stripping all conversational natural language, and (iv) preserving every mathematical definition and theorem signature verbatim without modifying the underlying formal semantics.

\begin{table}[H]
	\centering
	\caption{
		\textbf{\methodOurs{} generation pipeline.}
		The sequential, role-isolated agents composing the generation chain. Each stage processes specialized mathematical components, ensuring clear translation contracts before generating the final compilable Lean~4 code.
	}
	\label{tab:pipeline-agents}
	\small
	\resizebox{\textwidth}{!}{
	\begin{tabular}{@{}p{0.14\linewidth}p{0.18\linewidth}p{0.24\linewidth}p{0.38\linewidth}@{}}
		\toprule
		\textbf{Agent} & \textbf{Input} & \textbf{Output} & \textbf{Description \& Core Role} \\
		\midrule
		Distiller &
		\SNL, \PNL? &
		Domain + \ac{IG} context &
		Identifies the mathematical domain and either extracts the \ac{IG} from the proof or reformulates the open query into a declarative statement. \\
		\addlinespace
		Preprocessor &
		Distiller output &
		\SDNF, local defs, hyps, $G$ &
		Standardizes the problem into a unified \ac{SDNF} statement (\SDNF) and decomposes it into assumptions, local definitions, and goals. \\
		\addlinespace
		Formalizer &
		\SDNF components &
		Lean \texttt{def}s + \texttt{theorem \dots{} sorry} &
		Maps assumptions and infrastructure into idiomatic Lean~4 statements, relying on \textit{Mathlib} concepts while avoiding trivialization. \\
		\addlinespace
		Formatter &
		Draft code &
		Standalone \texttt{.lean} file &
		Organizes namespaces, imports, and scaffolding into a compilable file without altering the underlying mathematical meaning. \\
		\bottomrule
	\end{tabular}
	}
\end{table}

\subsection{Correction Loop Protocol}
\label[appendix]{app:loop-mechanics}

After the generation chain (decomposed or monolithic) produces an initial candidate statement, every multi-round variant enters the same iterative refinement skeleton for up to $T$ rounds.
Each round evaluates the current candidate under the loop's exit gate; if the gate fails and the round budget remains, a Corrector receives a structured briefing, emits a revised standalone Lean~4 file, and the candidate is re-verified.
All Correctors share a fixed output contract: emit a complete Lean~4 file beginning with \lean{import Mathlib}; terminate the main claim at \lean{:= by sorry} (or preserve an existing \lean{proof\_wanted} form); and forbid proof tactics, helper lemmas, \lean{admit}, or solving the underlying mathematics.
The two instantiations below differ only in the exit gate and in which signals populate the Corrector briefing.

Both the syntax-only and dual-signal loops share two mechanisms that stabilize iterative repair:

\paragraph{Syntactic Error De-cascading.}
In Lean~4, compiler diagnostics frequently suffer from extreme cascading: an early typo or minor type error in an initial definition can trigger dozens of downstream errors throughout the theorem header. Supplying an uncurated stream of 30+ compiler errors overwhelms language models, inducing destructive over-correction on otherwise sound code. To stabilize repair, the Lean Verifier extracts the first $k=5$ compiler errors to retain context, but explicitly instructs the Corrector to prioritize resolving the very first error ($e_1$). This top-down stabilization prevents chaotic thrashing across iterations.

\paragraph{Oscillation Mitigation via Prior Failed Attempts.}
Iterative repair loops can suffer from cyclic oscillation, where the model alternates between two conflicting fixes across successive rounds. To prevent this, the Corrector briefing dynamically maintains a \emph{Prior Failed Attempts} history log containing the statements attempted in previous rounds and their failure reasons. This history acts as a negative memory constraint, forcing the Corrector to explore alternative Mathlib abstractions rather than repeating previously rejected patterns.

\subsection{Syntax-Only Repair Loop}
\label[appendix]{app:syntax-loop}

The ablation variants \methodSyntaxLoop{} and \methodMonoSyntaxLoop{} isolate the contribution of compiler feedback without semantic gating.
After the same initial draft as their single-pass counterparts (decomposed or monolithic), they enter the shared repair skeleton above for up to $T$ rounds.
A dedicated syntax Corrector receives the informal statement, the current Lean draft, and the verifier report, then emits a revised signature file under the shared Corrector contract.
The loop exits successfully as soon as \metricCompile{} holds, or when the round budget~$T$ is exhausted.
Because termination depends solely on type-checking, a candidate may leave the loop as a compiling but mathematically unfaithful statement---the failure mode measured when comparing syntax-only rows to \methodOurs{} in \Cref{sec:experiments}.

\subsection{Dual-Signal Semantic Correction Loop}

When the initial draft is executed in the Lean~4 environment, \methodOurs{} evaluates it with both an objective compiler check and the in-loop semantic Rater.
A candidate exits successfully only when the joint gate \metricCompileAndOursSM{} is satisfied; otherwise the Corrector is invoked for up to $T$ rounds under the shared skeleton above.
In standard compile-only repair, a large language model (\acs{LLM}) often restores well-typedness by deleting or trivializing hypotheses, destroying the intended meaning.
To prevent this, the Corrector receives a structured bipartite diagnosis at every round:
\begin{enumerate}[nosep, leftmargin=*]
	\item \textbf{Syntactic feedback:} The first $k$ compiler error diagnostics (line, column, and message) from the Lean~4 verifier.
	\item \textbf{Semantic feedback:} Failing dimensions, scores, rationales, and localized mismatch snippets from the in-loop \textbf{Rater}, together with a Rater-proposed repair draft.
\end{enumerate}
The Corrector is instructed to prioritize \emph{semantic match over compilation}: a syntactically flawed but semantically faithful draft is treated as a productive intermediate state, whereas a compiling but mathematically incorrect statement must not be preserved.
The loop terminates successfully only when \metricCompileAndOursSM{} holds.

\paragraph{Pre-compilation of Semantic Drafts.}
During semantic evaluation, the Rater scores the candidate statement across the different gating dimensions and proposes an alternative draft designed to repair identified semantic mismatches. Instead of naively passing this draft to the Corrector, \methodOurs{} first compiles the Rater's draft in Lean~4. When the Corrector is invoked, it receives a synchronized bipartite diagnosis:
\begin{enumerate}[nosep, leftmargin=*]
	\item The compilation status and exact compiler diagnostics of its own previous attempt.
	\item The Rater's multidimensional semantic critique, the Rater's proposed draft, and the verified compilation status of that proposed draft.
\end{enumerate}
This mechanism prevents the Corrector from blindly adopting drafts that fix semantics but break syntax, forcing it to synthesize a solution that simultaneously satisfies both Lean's type checker and the semantic intent of the original mathematical query.

\paragraph{Anti-Solver and Anti-Injection Guard.}
In zero-shot and answer-agnostic settings, a naive rater may occasionally suggest ``fixing'' an existential goal by calculating the numerical answer and setting the variable to a constant or placeholder (e.g., suggesting $= 42$ or an uninstantiated constant $X$). To protect benchmark integrity, the Corrector is governed by a strict \emph{Anti-Solver Rule} and an \emph{Anti-Injection Escape Clause}. The agent is explicitly forbidden from performing calculations, solving the mathematical problem, or injecting concrete witnesses to silence type errors. If a proposed rater correction introduces an injected answer or an undefined placeholder, the Corrector is instructed to reject that modification and maintain the open-ended existential formulation (e.g., $\exists! N, \dots$).

\subsection{Compositional Generation Walkthrough}
\label[appendix]{app:walkthrough}

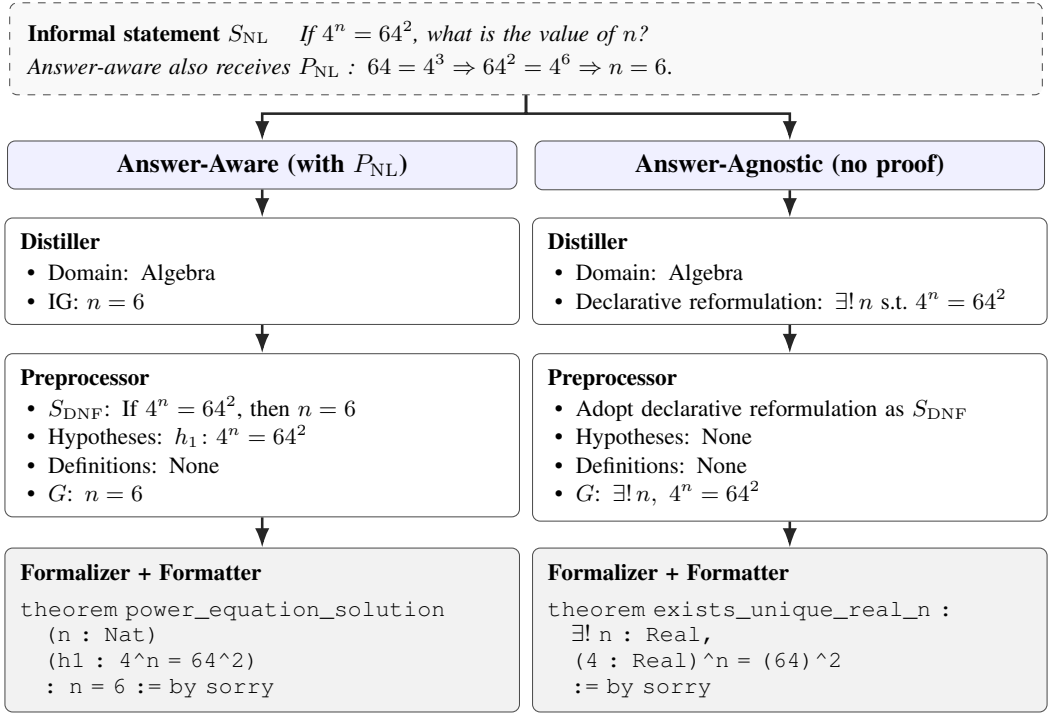
\begin{figure}[!h]
	\centering
	\resizebox{\textwidth}{!}{
		\begin{tikzpicture}[
			font=\footnotesize,
			>=Latex,
			x=1cm, y=1cm,
			box/.style={
				draw=black!70, rounded corners=3pt, align=left,
				inner sep=6pt, fill=white, text width=6.6cm,
				anchor=north
			},
			agent/.style={
				draw=black!70, fill=blue!6, rounded corners=3pt,
				align=center, inner sep=5pt, font=\bfseries,
				text width=6.6cm, anchor=north
			},
			inputbox/.style={
				draw=black!70, dashed, rounded corners=3pt, align=left,
				inner sep=7pt, fill=gray!5, text width=13.6cm
			},
			leanbox/.style={
				draw=black!70, rounded corners=3pt, align=left,
				inner sep=6pt, fill=gray!10, text width=6.6cm,
				anchor=north
			},
			arr/.style={->, line width=1pt, draw=black!85}
			]

			\node[inputbox] (inl) at (0,0) {
				\textbf{Informal statement}~\SNL\quad
				\emph{If $4^{n}=64^{2}$, what is the value of $n$?}\\[3pt]
				\textit{Answer-aware also receives}~\PNL\textit{:}
				$64=4^{3} \Rightarrow 64^{2}=4^{6} \Rightarrow n=6$.
			};

			\def\xL{-3.6}
			\def\xR{3.6}

			\def\yHead{-1.25}
			\def\yDist{-2.3}
			\def\yPrep{-4.15}
			\def\yLean{-6.8}

			\node[agent] (hL) at (\xL,\yHead) {Answer-Aware (with~\PNL)};
			\node[agent] (hR) at (\xR,\yHead) {Answer-Agnostic (no proof)};

			\node[box] (dL) at (\xL,\yDist) {
				\textbf{Distiller}\\[2pt]
				\begin{itemize}[nosep,leftmargin=1.2em,itemsep=1pt,topsep=2pt,parsep=0pt]
					\item Domain: Algebra
					\item \ac{IG}: $n = 6$
				\end{itemize}
			};
			\node[box] (dR) at (\xR,\yDist) {
				\textbf{Distiller}\\[2pt]
				\begin{itemize}[nosep,leftmargin=1.2em,itemsep=1pt,topsep=2pt,parsep=0pt]
					\item Domain: Algebra
					\item Declarative reformulation:
					$\exists!\, n$ s.t.\ $4^{n}=64^{2}$
				\end{itemize}
			};

			\node[box] (pL) at (\xL,\yPrep) {
				\textbf{Preprocessor}\\[2pt]
				\begin{itemize}[nosep,leftmargin=1.2em,itemsep=1pt,topsep=2pt,parsep=0pt]
					\item \SDNF: If $4^{n}=64^{2}$, then $n=6$
					\item Hypotheses: $h_1\colon 4^{n}=64^{2}$
					\item Definitions: None
					\item $G$: $n = 6$
				\end{itemize}
			};
			\node[box] (pR) at (\xR,\yPrep) {
				\textbf{Preprocessor}\\[2pt]
				\begin{itemize}[nosep,leftmargin=1.2em,itemsep=1pt,topsep=2pt,parsep=0pt]
					\item Adopt declarative reformulation as \SDNF
					\item Hypotheses: None
					\item Definitions: None
					\item $G$: $\exists!\, n,\ 4^{n}=64^{2}$
				\end{itemize}
			};

			\node[leanbox] (fL) at (\xL,\yLean) {
				\textbf{Formalizer + Formatter}\\[4pt]
				{\ttfamily
					theorem power\_equation\_solution\\
					\quad(n : Nat)\\
					\quad(h1 : 4\textasciicircum{}n = 64\textasciicircum{}2)\\
					\quad: n = 6 := by sorry}
			};
			\node[leanbox] (fR) at (\xR,\yLean) {
				\textbf{Formalizer + Formatter}\\[4pt]
				{\ttfamily
					theorem exists\_unique\_real\_n :\\
					\quad$\exists!$ n : Real,\\
					\quad(4 : Real)\textasciicircum{}n = (64)\textasciicircum{}2\\
					\quad:= by sorry}
			};

			\draw[arr] (inl.south) -- ++(0,-0.25) -| (hL.north);
			\draw[arr] (inl.south) -- ++(0,-0.25) -| (hR.north);
			\draw[arr] (hL.south) -- (dL.north);
			\draw[arr] (hR.south) -- (dR.north);
			\draw[arr] (dL.south) -- (pL.north);
			\draw[arr] (dR.south) -- (pR.north);
			\draw[arr] (pL.south) -- (fL.north);
			\draw[arr] (pR.south) -- (fR.north);
			
		\end{tikzpicture}
	}
	\caption{
		\textbf{Compositional generation walkthrough} on \datasetOmni{} \#043.
		The same informal question is processed under both regimes.
		With~\PNL, the Distiller extracts an \ac{IG} ($n=6$) and the Preprocessor
		builds a \ac{SDNF} claim with hypothesis $h_1$ and goal $G\colon n=6$.
		Without a proof, the Distiller commits to a declarative reformulation
		($\exists!\, n$), which flows through to an existential Lean~4 goal.
	}
	\label{fig:pipeline-example}
\end{figure}

\Cref{fig:pipeline-example} illustrates the complete execution trace of the \methodOurs{} pipeline on problem \datasetOmni{} \#043 across both operational regimes:
\begin{itemize}[nosep, leftmargin=*]
	\item \textbf{Answer-Aware Regime (Guided by $\PNL$):} The Distiller analyzes the informal proof $64 = 4^3 \Rightarrow 64^2 = 4^6 \Rightarrow n = 6$ and extracts the Inferred Goal $n = 6$. The Preprocessor generates \SDNF: \textit{``If $4^n = 64^2$, then $n = 6$''}, binds hypothesis $h_1\colon 4^n = 64^2$, and sets goal $G\colon n = 6$. The Formalizer produces an explicit verification theorem asserting $n = 6$.
	\item \textbf{Answer-Agnostic Regime (Zero-Shot without Proof):} Deprived of an answer oracle, the Distiller refuses to compute the exponent. It reformulates the open query into an objective unique existential claim ($\exists!\, n$). The Preprocessor binds this into goal $G\colon \exists!\, n,\ 4^n = 64^2$, and the Formalizer generates an existential theorem in Lean~4, delegating the computation of $n$ to downstream provers.
\end{itemize}

\begin{revblock}
\subsection{Architectural Comparison with Prior Autoformalization Frameworks}
\label[appendix]{app:prior-frameworks}

To contextualize the architectural design of \methodOurs{}, \Cref{tab:prior-frameworks} contrasts our four-stage multi-agent decomposition and direct Lean~4 verification loop against three representative multi-stage or benchmark-curation frameworks: \textsf{FIMO}~\citep{liu2023fimo}, \textsf{KELPS}~\citep{zhang2025kelps}, and \textsf{LoC-Decomp}~\citep{shi2026locdecomp}.

\paragraph{1. Multi-Stage Agentic Isolation vs.\ Syntactic Templates and Closed IRs.}
\textsf{FIMO}~\citep{liu2023fimo} applies no agentic decomposition: it converts open-ended Algebra and Number Theory problems into closed proof targets by manually appending the official contest answer to the prompt prior to monolithic GPT-4 translation, compiler feedback, and human expert post-editing.
\textsf{KELPS}~\citep{zhang2025kelps} introduces a two-stage neuro-symbolic pipeline that parses informal text into an assertional-logic intermediate representation (\emph{Knowledge Equations}) restricted to a closed ontology of $40$ concepts and $180$ operators, before applying deterministic ANTLR4 transpilation rules that emit dual-\lean{sorry} templates (\lean{\{x | ...\} = sorry := by sorry}) on solve-type queries.
\textsf{LoC-Decomp}~\citep{shi2026locdecomp} generates Lean~4 code in a single LLM call (\textsf{FormalTrans}) guided by an in-prompt four-marker template (\lean{structure MySystem} plus auxiliary definitions, constraints, and \lean{def problem\_statement}), slicing those markers post hoc for segmented natural-language back-translation.
By contrast, \methodOurs{} isolates \emph{distinct stage roles and information boundaries} across four dedicated stages (Distiller $\to$ Preprocessor $\to$ Formalizer $\to$ Formatter). Enforcing this boundary before Lean synthesis prevents the model from entering contest-solver mode (reducing answer leakage from \qty{63.0}{\percent} under single-call anti-solver prompting to \qty{1.0}{\percent} at $T{=}1$; \Cref{app:anti-solver}) while emitting unwrapped, idiomatic Mathlib \lean{theorem} declarations over arbitrary Olympiad geometry, combinatorics, and undergraduate domains.

\paragraph{2. Target Representation Conventions: Open Predicates vs.\ Closed Single-\texttt{sorry} Theorems.}
Prior frameworks also adopt distinct target representations for open-ended questions. \textsf{LoC-Decomp} formalizes solve-type problems as open predicate specifications \lean{def problem\_statement (solution : $\alpha$) : Prop} (of type $\alpha \to \texttt{Prop}$) paired with a \lean{MySystem} parameter bundle, whereas \methodOurs{} targets closed \lean{Prop} declarations under the read-only single-\lean{sorry} theorem convention expected by standard automated theorem provers. Targeting closed single-\lean{sorry} theorems imposes distinct type-theoretic constraints: in Lean~4, a \lean{theorem} declaration requires a closed \lean{Prop} return type, and placing \lean{(solution : $\alpha$)} in a \lean{theorem} binder universally quantifies $\forall\,\text{solution}$ rather than requesting a witness. Translating open queries into closed theorems without an answer oracle therefore directly confronts our \emph{Feasibility Trilemma} (\Cref{sec:limits})---where naive existential wrapping ($\exists s,\, P(s)$) admits non-constructive shortcuts such as the \emph{Triviality Trap} (\lean{use \{x | P x\}; rfl}) or \emph{Polarity Cheat} (\lean{Classical.em}), motivating Type-Theoretic Locks (\Cref{app:type_locks})---and requires scoping witness quantification inside problem parameters ($\forall\,\text{sys}\,\exists s$ rather than $\exists s\,\forall\,\text{sys}$) whenever the witness depends on free system variables.

\paragraph{3. Direct Lean~4 Semantic Gating vs.\ Natural-Language Back-Translation.}
During verification, \textsf{LoC-Decomp} slices Lean~4 code at template markers, back-translates each chunk independently into informal prose (\textsf{BackTrans}), merges the resulting text, and alternates between natural-language consistency loops (\textsf{ASCC-R}) and compiler repair loops (\textsf{CpC-R}). Evaluating fidelity through segmented natural-language back-translation introduces an inherently lossy intermediate channel: formal type-theoretic semantics can be obscured when mapped back to prose---for instance, truncating natural-number division \lean{(1 : Nat) / 2 = 0} reads benignly as ``half of''---while translating code chunks in isolation can lose cross-segment binder and quantifier dependencies or penalize valid formalizations whose back-translated phrasing diverges stylistically from the original problem text. \methodOurs{} instead audits the complete Lean~4 declaration directly against multi-dimensional formal rubrics (\metricOursSM{}) and pre-compiles rater-proposed repair drafts within the same turn (\Cref{app:loop-mechanics}), resolving syntactic and semantic faults jointly without natural-language round-tripping.

\begin{table}[H]
	\centering
	\caption{\rev{\textbf{Architectural comparison with prior autoformalization frameworks.}}}
	\label{tab:prior-frameworks}
	\footnotesize
	\setlength{\tabcolsep}{4.5pt}
	{\color{revblue}
	\begin{tabular}{@{} >{\raggedright\arraybackslash}p{0.15\textwidth} >{\raggedright\arraybackslash}p{0.17\textwidth} >{\raggedright\arraybackslash}p{0.18\textwidth} >{\raggedright\arraybackslash}p{0.20\textwidth} >{\raggedright\arraybackslash}p{0.20\textwidth} @{}}
		\toprule
		\textbf{Dimension} & \textbf{\textsf{FIMO}}~\citep{liu2023fimo} & \textbf{\textsf{KELPS}}~\citep{zhang2025kelps} & \textbf{\textsf{LoC-Decomp}}~\citep{shi2026locdecomp} & \textbf{\methodOurs{} (Ours)} \\
		\midrule
		\textbf{Decomposition} & None (monolithic single-pass LLM) & Two-stage neural KE DSL + deterministic ANTLR4 rules & Single-call LLM with 4-marker \texttt{MySystem} prompt template & 4-stage multi-agent role \& information isolation \\
		\addlinespace[4pt]
		\textbf{Question / Solve Target} & Official contest answer manually appended in prompt & Dual-\texttt{sorry} set equality (\texttt{\{x | ...\} = sorry := by sorry}) & Open \texttt{def} predicate \texttt{(s : $\alpha$) : Prop} ($\alpha \to \texttt{Prop}$) & Closed \texttt{Prop} \texttt{theorem} or Type-Theoretic Locks ($1\times$ \texttt{sorry}) \\
		\addlinespace[4pt]
		\textbf{Semantic Verification} & Manual post-hoc human expert inspection & Offline post-generation filtering only & Segmented NL \texttt{BackTrans} + prose consistency (\texttt{ASCC}) & Direct Lean~4 multi-dimensional rubric gate (\metricOursSM{}) \\
		\addlinespace[4pt]
		\textbf{Refinement Loop} & Compiler feedback + manual human post-editing & Symbolic grammar transpilation (no semantic loop) & Alternating NL semantic (\texttt{ASCC-R}) \& syntax (\texttt{CpC-R}) loops & Simultaneous dual-signal loop + pre-compiled repair draft \\
		\addlinespace[4pt]
		\textbf{Domain Scope} & Olympiad Algebra \& Number Theory & Closed ontology ($40$ concepts, $180$ operators) & Competition math \& \textsc{ProofNet} (\texttt{MySystem} wrapper) & Unwrapped native Mathlib (Geo, Combi, Alg, NT, UG) \\
		\bottomrule
	\end{tabular}%
	}
\end{table}
\end{revblock}

%% file: appendix_setup.tex
\section{Experimental Setup, Datasets, and Reproducibility}
\label[appendix]{app:setup}
\label[appendix]{app:dataset-details}

In this section, we provide comprehensive details on the benchmark datasets, the language model serving infrastructure, the evaluated model backbones, our compute environment constraints, and the full configuration matrix across all experimental ablations.

\subsection{Dataset Details and Curation}
\label[appendix]{app:datasets}

To test autoformalization across diverse problem types and formalization regimes, our study evaluates on three benchmarks: \datasetOmni{}, \datasetImoFormalized{}, and \datasetImo{}. Both IMO suites are derived from a comprehensive snapshot of the community archive taken in July 2026.

\paragraph{\datasetOmni{} ($N=\datasetOmniN$).}
\datasetOmni{}~\citep{gao2025omni} is an olympiad-level competition benchmark spanning contests such as AMC 10/12, AIME, USAMO, and Putnam. We evaluate on the exact 300-problem subset benchmarked by Goedel-Prover-V2~\citep{lin2025goedelproverv2} to ensure direct comparability.
\begin{itemize}[nosep, leftmargin=*]
	\item \textbf{Curation Protocol:} Extracted directly from the standardized 300-problem evaluation split of \citet{lin2025goedelproverv2}.
	\item \textbf{Problem Texture:} \emph{Question-type} \qty{98.7}{\percent}; \emph{Assertion-type} \qty{1.3}{\percent}. Question-type statements require calculating or identifying unknown extremal values, witnesses, or configurations.
	\item \textbf{Domain Distribution:} From the dataset metadata primary subject tags: Algebra (\qty{62.7}{\percent}), Number Theory (\qty{29.0}{\percent}), Combinatorics (\qty{6.3}{\percent}), and Precalculus (\qty{2.0}{\percent}). This 300-subset contains no Geometry-tagged problems.
\end{itemize}

\paragraph{\datasetImoFormalized{} ($N=\datasetImoFormalizedN$).}
This benchmark comprises all historical International Mathematical Olympiad (IMO) problems from 1959 to 2024 that possess official, human-verified formalizations in Lean~4.
\begin{itemize}[nosep, leftmargin=*]
	\item \textbf{Curation Protocol:} Extracted from a snapshot of the \href{https://github.com/dwrensha/compfiles}{\texttt{compfiles}} repository~\citep{renshaw2024compfiles} taken in July 2026. It represents the fraction of historical IMO problems ($223 / 398$) already formalized by the Lean community, with informal statements extracted from module docstrings and formal declarations parsed into standalone targets.
	\item \textbf{Problem Texture:} Nearly balanced: \emph{Question-type} \qty{51.6}{\percent}; \emph{Assertion-type} \qty{48.4}{\percent}.
	\item \textbf{Domain Distribution:} Distiller-assigned primary domains, mapped to the four canonical IMO disciplines (analysis / functional-equation labels folded into Algebra): Number Theory (\qty{33.2}{\percent}), Algebra (\qty{33.6}{\percent}), Combinatorics (\qty{26.5}{\percent}), and Geometry (\qty{6.7}{\percent}).
\end{itemize}

\paragraph{\datasetImo{} ($N=\datasetImoN$).}
This benchmark comprises the complementary frontier of historical IMO problems that have never been formalized in Lean~4.
\begin{itemize}[nosep, leftmargin=*]
	\item \textbf{Curation Protocol:} Derived from the same July 2026 \texttt{compfiles} snapshot by strictly selecting the complement fraction ($175 / 398$) that lacked official human formalizations. Problem texts and informal solution notes were curated from the AI-MO olympiads compendium and official IMO archives.
	\item \textbf{Problem Texture:} \emph{Assertion-type} \qty{64.0}{\percent}; \emph{Question-type} \qty{36.0}{\percent}.
	\item \textbf{Domain Distribution:} Distiller-assigned primary domains under the same four-discipline mapping: Geometry (\qty{69.1}{\percent}), Combinatorics (\qty{24.6}{\percent}), Algebra (\qty{3.4}{\percent}), and Number Theory (\qty{2.9}{\percent}). The Geometry skew reflects which historical IMO problems remain unformalized in Lean~4.
\end{itemize}

\subsection{LLM Serving, Models, and Infrastructure}
\label[appendix]{app:llm-settings}
\label[appendix]{app:models}

All language model inference is executed using the high-performance \texttt{vLLM} engine. Each model is served on dedicated GPU instances with a tensor parallelism degree of 1, utilizing FP8 quantized weights and dynamic KV caching to maximize token throughput during iterative multi-agent correction loops.

\paragraph{Evaluated Models and Pipeline Roles.}
Our experimental framework strictly decouples generation, in-loop rating, and post-hoc evaluation to eliminate circular self-evaluation bias:
\begin{enumerate}[nosep, leftmargin=*]
	\item \textbf{Generation \& In-Loop Rating (\modelQwenLong):} We deploy \texttt{Qwen/Qwen3.6-27B-FP8} as the core workhorse model. It powers all four stages of the generation chain (Distiller, Preprocessor, Formalizer, Formatter) as well as the in-loop semantic Rater and Statement Corrector.
	\item \textbf{Independent Post-Hoc Evaluator (\modelGemmaLong):} To ensure an unbiased, external evaluation of semantic fidelity, all post-hoc consensus judging (\metricGoedelSM) and blind pairwise win-rate adjudications (\metricPWR) are performed using \texttt{RedHatAI/gemma-4-31B-it-FP8-Dynamic}.
	\item \textbf{Domain-Specialized Baseline (\modelGoedelLong):} We evaluate against \texttt{Goedel-LM/Goedel-Formalizer-V2-32B}~\citep{lin2025goedelproverv2}, an open-weights model explicitly fine-tuned for Lean~4 statement autoformalization.
\end{enumerate}

\Cref{tab:llm-settings} details the decoding hyperparameters, sampling bounds, and generation settings across each model role.

\begin{table}[H]
	\centering
	\caption{\textbf{LLM decoding and sampling hyperparameters.}}
	\label{tab:llm-settings}
	\small
	\begin{tabular}{@{}llccr@{}}
		\toprule
		\textbf{Model} & \textbf{Role} & \textbf{Temp.} & \textbf{Max tokens} & \textbf{Retries} \\
		\midrule
		\modelQwenLong{}
		& Pipeline generation + in-loop rater
		& 0.6 & 16{,}384 & 2 \\
		\modelGoedelLong{}
		& Fine-tuned baseline (\methodGoedel)
		& 0.7 & 16{,}384 & 1 \\
		\modelGemmaLong{}
		& Post-hoc evaluation (\metricGoedelSM{} / \metricPWR{})
		& 0.6 & 16{,}384 & 2 \\
		\bottomrule
	\end{tabular}
\end{table}

\paragraph{Note on Model Backbones and Qwen 3.8.}
All primary experiments, benchmark sweeps, and ablation ladders were established using \modelQwenLong{}, which served as our stable workhorse throughout development. \modelQwenThreeEight{} was released late in our experimental cycle. While its enhanced capabilities yield substantial performance gains in our scaling ablation (\Cref{tab:omni-backend-scaling}), its extended reasoning traces noticeably increased per-step latency. Given the multi-agent nature and multi-round repair loops of our pipeline, re-executing the full experimental suite across all benchmarks was impractical within our project timeline. We therefore maintain \modelQwenLong{} as the consistent backbone across all primary comparative tables.

\paragraph{Compute Environment and Model Access.}
\ificlrfinal
All experiments and evaluations were conducted on an on-premise computing cluster at the Huawei Lagrange Mathematics and Computing Research Center. Due to internal corporate compliance and data governance policies, our environment does not have access to external proprietary APIs or closed-source commercial models. Consequently, our pipeline and benchmark evaluations rely exclusively on open-weights models deployed on local infrastructure. Beyond adhering to corporate requirements, this ensures full experimental reproducibility and demonstrates that \methodOurs{} delivers strong autoformalization performance without relying on proprietary black-box models.
\else
All experiments and evaluations were conducted on an on-premise computing cluster equipped with dedicated GPU nodes. To prioritize scientific reproducibility, eliminate dependency on closed commercial APIs with volatile endpoints, and ensure strict data governance, our framework relies exclusively on publicly available, open-weights models deployed locally. This setup ensures that all benchmarks, latency profiles, and generated formal artifacts are fully deterministic, auditable, and reproducible by the broader research community without reliance on proprietary black-box services.
\fi

\subsection{Method Variants and Ablation Configurations}
\label[appendix]{app:methods}

To systematically isolate the individual contributions of pipeline decomposition, syntactic error feedback, and semantic rating feedback, we benchmark six primary configurations across two operational regimes, detailed in \Cref{tab:methods}.
Evaluating these variants on an identical open-weights backbone (\modelQwenLong{}) allows us to causally decouple the effects of structural decomposition and feedback signals without confounding differences in base model capability or pretraining data.
For all iterative methods, the maximum repair budget is set to $T=5$ rounds on \datasetOmni{} and $T=10$ rounds on \datasetImo{} \rev{under matched iteration caps (with full generation token accounting in \Cref{app:token-accounting})}.

\paragraph{Methodological Framing and \rev{Iteration Budget} Rationale.}
Autoformalization literature features various systems with disparate base models and training setups (e.g., Process-Driven Autoformalization~\citep{lu2024process} and ReForm~\citep{chen2026reform}).
Direct cross-system benchmarking against published checkpoints inevitably confounds algorithmic framework merits with base model capacity and pretraining exposure.
To provide rigorous scientific attribution, our evaluation strategy is two-fold:
\begin{enumerate}[leftmargin=*, nosep]
	\item \textbf{Controlled Mechanism Isolation:} We benchmark the foundational mechanisms of compiler-only refinement (\methodSyntaxLoop{}, isolating the compiler-feedback loop of \citealp{lu2024process}) and all-in-one reflective self-critique (\methodMonoSemanticLoop{}, capturing the reflective paradigm of \citealp{chen2026reform}) on an \emph{identical} open-weights backbone (\modelQwenLong{}) under \emph{identical} iteration budgets ($T=5$ and $T=10$), isolating the precise value added by four-stage decomposition and dual-signal verification \rev{under matched iteration budgets}.
	\item \textbf{Fine-Tuning Paradigm Comparison:} We include \methodGoedelLong{} as an external anchor representing the specialized supervised fine-tuning paradigm. While this baseline uses single-pass generation ($T=1$), it embodies a specialized 32B model fine-tuned via expert iteration on 50{,}000 Claude-distilled formalization traces; comparing against it highlights the trade-off between static parameter fine-tuning and zero-training test-time neurosymbolic scaffolding.
\end{enumerate}

\paragraph{Overview of Evaluated Configurations.}
Our benchmark suite spans the following method variants:
\begin{itemize}[leftmargin=*]
	\item \textbf{Single-Pass Forward Baselines ($T=1$):}
	\begin{itemize}[nosep, leftmargin=*]
		\item \methodMonolithic{}: Translates the informal problem into a formal Lean~4 statement in a single forward pass without intermediate reasoning scaffolding. The candidate is then normalized by a statement Formatter (imports and file assembly) before compilation; the same formatting pass is used for all Qwen configurations.
		\item \methodDecomposition{}: Executes our four-stage feedforward generation chain (Distiller $\to$ Preprocessor $\to$ Formalizer $\to$ Formatter) without entering a repair loop.
		
	\end{itemize}
	\item \textbf{Compiler-Only Feedback Loops ($T=5$ / $T=10$):}
	\begin{itemize}[nosep, leftmargin=*]
		\item \methodMonoSyntaxLoop{}: An iterative loop wrapping the monolithic generator, utilizing raw Lean~4 compiler diagnostics for syntactic self-repair.
		\item \methodSyntaxLoop{}: An iterative loop wrapping our four-stage pipeline, providing compiler error diagnostics to the Corrector agent while omitting semantic critique.
	\end{itemize}
	\item \textbf{Dual-Signal Semantic Refinement Loops ($T=5$ / $T=10$):}
	\begin{itemize}[nosep, leftmargin=*]
		\item \methodMonoSemanticLoop{}: An all-in-one reflective loop where the monolithic candidate is evaluated by both the compiler and the in-loop semantic Rater (\metricOursSM{}), providing dual-signal feedback to revise the statement.
		\item \textbf{\methodOurs{}} (\methodSemanticLoop{}): Our complete framework, combining the four-stage generation pipeline with dual-signal syntactic and semantic feedback.
	\end{itemize}
	\item \textbf{External Supervised Baseline ($T=1$):}
	\begin{itemize}[nosep, leftmargin=*]
		\item \methodGoedel{} (\modelGoedelLong{}): Evaluated on its raw single-pass output under its official thinking-prompt format, without an additional formatting or repair stage, to calibrate performance against specialized supervised learning.
	\end{itemize}
\end{itemize}

\begin{table}[H]
	\centering
	\caption{\textbf{Overview of evaluated methods and ablation variants.} Iterative variants are evaluated under matching iteration budgets ($T=5$ on \datasetOmni{}, $T=10$ on \datasetImo{}) on a shared base model\rev{; full generation token accounting is reported in \Cref{app:token-accounting}}.}
	\label{tab:methods}
	\small
	\resizebox{\textwidth}{!}{
		\begin{tabular}{@{}l l l l c@{}}
			\toprule
			\textbf{Method} & \textbf{Base Model} & \textbf{Architecture} & \textbf{Feedback Type} & \textbf{Rounds ($T$)} \\
			\midrule
			\multicolumn{5}{@{}l}{\textit{\textbf{Monolithic Configurations}}} \\
			\midrule
			\methodMonolithic & \modelQwenLong & Monolithic & None & 1 \\
			\methodMonoSyntaxLoop & \modelQwenLong & Monolithic & \metricCompile & 5--10 \\
			\methodMonoSemanticLoop & \modelQwenLong & Monolithic & \metricCompile + \metricOursSM & 5--10 \\
			\midrule
			\multicolumn{5}{@{}l}{\textit{\textbf{Decomposed Configurations (Ours)}}} \\
			\midrule
			\methodDecomposition & \modelQwenLong & Decomposed & None & 1 \\
			\methodSyntaxLoop & \modelQwenLong & Decomposed & \metricCompile & 5--10 \\
			\textbf{\methodOurs{}} (\methodSemanticLoop) & \modelQwenLong & Decomposed & \metricCompile + \metricOursSM & 5--10 \\
			\midrule
			\multicolumn{5}{@{}l}{\textit{\textbf{External Domain Baseline}}} \\
			\midrule
			\methodGoedel & \modelGoedelLong & Monolithic (Thinking) & None  & 1 \\
			\bottomrule
		\end{tabular}
	}
\end{table}

\subsection{Evaluation Metrics and Integrity Audits}
\label[appendix]{app:metrics}
\label[appendix]{app:audit-agents}

Standard autoformalization evaluations typically report raw compilation rates (\metricCompile{}), implicitly assuming that syntactically valid code implies mathematical correctness. In reality, models frequently generate type-correct formalizations that are mathematically vacuous, structurally altered, or unfaithful to the informal source. To establish an honest evaluation standard, our protocol decouples syntactic compilation from post-hoc semantic validation, and introduces strict automated audits for benchmark integrity.

\paragraph{Compilation Environment (\metricCompile{}).}
All Lean~4 compilation checks are executed in isolated worker processes managed by our verification engine. Every candidate file imports Mathlib and is verified against a pinned Lean~4 toolchain. A formalization passes \metricCompile{} if the environment returns an exit code of 0 without diagnostic type errors. Proof bodies are omitted via \lean{:= by sorry}. Statements that trigger compiler timeouts ($>$\qty{30}{\second}) or memory exhaustion are treated as compilation failures.

We pin Lean~4 to \texttt{v4.29.0-rc2} with a fixed Mathlib revision via the project lockfile; every candidate imports Mathlib and is checked in this environment.

\paragraph{The Single-Sorry Invariant and Cheat Detection.}
As introduced in \Cref{sec:sage-verification}, a valid statement must respect the \textbf{single-sorry contract}. Permitting multiple placeholders exposes benchmarks to two severe vulnerabilities:
\begin{enumerate}[nosep, leftmargin=*]
	\item \emph{Definition Stubbing:} Models can declare unresolved constant definitions stubbed with a placeholder (e.g., \lean{def k : Nat := sorry}), and subsequently state a trivial theorem over \lean{k}. While syntactically valid, any proof over an uninstantiated definition is vacuous.
	\item \emph{Hypothesis and Lemma Evasion:} Models frequently decompose hard theorems into intermediate helper lemmas closed by \lean{sorry} (e.g., \lean{lemma helper : ... := by sorry}), effectively assuming the problem's deepest mathematical barrier rather than stating a self-contained target.
\end{enumerate}
Our audit strips comments and string literals from the Lean source, then counts unresolved placeholder tokens (\lean{sorry}, \lean{admit}, \lean{proof\_wanted}, and related forms). A prover-ready candidate must contain exactly one terminating placeholder on the primary target declaration (\lean{theorem}, or \lean{def} for Type locks). Any compiling candidate with more than one placeholder, or with placeholders inside intermediate definitions or helper lemmas that are \emph{not} the primary target, is disqualified as \emph{Multiple Sorries}.

\paragraph{Answer Leakage Audit Protocol.}
When evaluating open-ended \emph{Question-type} problems under the answer-agnostic (without-proof) regime, models are tasked with formalizing the question itself, not computing the solution. However, monolithic models frequently shortcut the translation by searching for the answer during generation and hardcoding the numeric result into the theorem signature (e.g., translating ``Find the minimum value of $f(x)$'' into \lean{theorem T : $\forall$ x, f x $\geq$ 42}).
We audit this with an independent LLM judge (\modelGemmaLong{}). Given the informal statement, the informal proof when available, and the generated Lean code, the auditor runs a fixed protocol:
\begin{enumerate}[nosep, leftmargin=*]
	\item \textbf{Texture classification.} Decide whether the informal problem is Assertion-type (self-contained claim) or Question-type (asks to compute, determine, or construct an object). Assertion-type problems are excluded from leakage penalties, even if their mathematics involves existentials.
	\item \textbf{Witness presence.} Check whether the formal statement embeds a concrete witness term or numeric value rather than an open encoding (e.g., $\exists$ / $\exists!$).
	\item \textbf{Match and type check.} If a witness is embedded, check whether it matches the ground-truth solution and whether its Lean type is compatible with what the informal problem asks for.
\end{enumerate}
In the answer-agnostic regime, any compiling Question-type formalization that embeds a concrete witness is flagged as \textbf{Answer Leakage}.
Ground-truth matching in step~3 is diagnostic only: the binary leakage flag does not require the witness to equal the dataset answer.

\begin{revblock}
\subsubsection{Ground-Truth Correctness of Leaked Witnesses}
\label[appendix]{app:leak-truth}
In the answer-agnostic setting, injecting a concrete witness violates the autoformalizer's contract even when the guess happens to be right: (i)~Lean~4 type-checks \lean{theorem T : x = 42 := by sorry} and \lean{theorem T : x = 43 := by sorry} identically, providing zero compiler signal when an unverified guess is wrong; (ii)~supplying the witness alters the mathematical task from open-ended \emph{discovery} to closed \emph{verification}; and (iii)~correct injections on public benchmarks can reflect pretraining memorization rather than faithful translation.
To quantify how often leaked guesses are empirically correct, \Cref{tab:leak-truth} reports the diagnostic step-3 solution match across compiling Question-type drafts that embed a witness on \datasetOmni{} (answer-agnostic, pass@1). Nearly half of compiling leaks in monolithic pipelines (\qty{46.2}{\percent} for \methodMonolithic{}, \qty{48.5}{\percent} for \methodMonoSemanticLoop{}) and \qty{70.7}{\percent} of \methodGoedel{} leaks assert a mathematically false conjecture; \methodOurs{} incurs only $8$ compiling leaks out of $300$ problems (\qty{2.7}{\percent}). Note that \Cref{tab:omni-goedel-reject} reports a distinct denominator: multi-hot rationale tags among drafts marked \emph{Inappropriate} by \metricGoedelSM{}, rather than shares of leaked drafts.

\begin{table}[H]
  \centering
  \caption{\rev{\textbf{Ground-truth accuracy of leaked witnesses on \datasetOmni{} (Answer-Agnostic, pass@1).} Step-3 solution audit over compiling Question-type drafts that embed a concrete witness ($N=\datasetOmniN$).}}
  \label{tab:leak-truth}
  \small
  {\color{revblue}
  \begin{tabular}{@{}l rrr r@{}}
    \toprule
    \textbf{Method} & \textbf{Compiling Leaks} & \textbf{Matches Solution} & \textbf{False Witness} & \textbf{False Share (\%)} \\
    \midrule
    \methodMonolithic & 158 & 85 (53.8\%) & 73 & 46.2 \\
    \methodMonoSemanticLoop & 202 & 104 (51.5\%) & 98 & 48.5 \\
    \methodGoedelLong & 123 & 36 (29.3\%) & 87 & 70.7 \\
    \methodOurs & 8 & 3 (37.5\%) & 5 & 62.5 \\
    \bottomrule
  \end{tabular}
  }
\end{table}
\end{revblock}

\paragraph{Post-Hoc Consensus Semantic Match (\metricGoedelSM{}).}
For comparability with prior work, we report post-hoc semantic match under the Goedel-Prover-V2 protocol~\citep[Appendix D]{lin2025goedelproverv2}: an independent \modelGemmaLong{} judge, queried repeatedly, with success only under unanimous \emph{Appropriate} consensus. Full consensus settings, prompt-bias analysis, and reject-reason taxonomy are in \Cref{app:goedel-bias}. Because that judge prefers explicit answer injection, leakage audits are essential when interpreting these scores.

\paragraph{Multi-Dimensional In-Loop Semantic Gate (\metricOursSM{}).}
Unlike a single post-hoc \emph{Appropriate}/\emph{Not Appropriate} prompt, our in-loop gate (\metricOursSM{}) evaluates formalizations with the multi-dimensional Rater from \Cref{sec:sage-verification}.
A candidate passes \metricOursSM{} iff all seven gated dimensions clear the cutoffs in \Cref{tab:sage-sm-gate}.
Auxiliary diagnostic scores (e.g., compilability and difficulty) inform repair suggestions but do not enter the binary gate. The gate steers the Statement Corrector during search and can also filter high-trust training statements post hoc.

\begin{table}[H]
  \centering
  \caption{\textbf{In-loop \metricOursSM{} gate.} Scales are those used by the Rater prompt; cutoffs are conjunctive (all must hold).}
  \label{tab:sage-sm-gate}
  \small
  \begin{tabular}{@{}lcc@{}}
    \toprule
    \textbf{Dimension} & \textbf{Scale} & \textbf{Cutoff} \\
    \midrule
    Semantic Match & $0$--$4$ & $\ge 4$ \\
    Missing & $0$--$3$ & $\le 1$ \\
    Extra & $0$--$4$ & $\le 1$ \\
    Wrong & $0$--$2$ & $= 0$ \\
    Exactness & $0$--$3$ & $\ge 1$ \\
    Naturality & $0$--$4$ & $\ge 3$ \\
    Descriptive Precision & $0$--$4$ & $\ge 3$ \\
    \bottomrule
  \end{tabular}
\end{table}

\paragraph{Blind Pairwise Win Rate (\metricPWR{}).}
To evaluate semantic fidelity without rigid prompt heuristics, we perform blind head-to-head pairwise comparisons between model outputs $\mathcal{A}$ and $\mathcal{B}$. An independent judge compares both formalizations against the informal source on a 5-point Likert scale: \emph{Much Worse} ($-2$), \emph{Worse} ($-1$), \emph{Tie} ($0$), \emph{Better} ($+1$), and \emph{Much Better} ($+2$). To eliminate positional bias, every pair is evaluated twice with swapped presentation orders: $(\mathcal{A}, \mathcal{B})$ and $(\mathcal{B}, \mathcal{A})$. Inconsistent orders are reconciled via tie assignment. We report preference shares over all comparisons (strict wins, ties, and losses); conditional on non-ties, the win rate is the share of decisive comparisons where a method is strictly preferred.

\paragraph{Joint Evaluation Criteria and Sample Aggregation.}
Our primary evaluation metric is the joint criterion $\metricCompileAndGoedelSM{}$, denoting candidate statements that are simultaneously well-typed in Lean~4 and unanimously approved by the post-hoc semantic judge. For multi-sample search ($\text{pass@}N$), a problem is counted as solved under a given metric if \emph{at least one} of the $N$ independent candidate generations satisfies the required criteria.

\subsection{Goedel Evaluation Protocol and Prompt Bias Analysis}
\label[appendix]{app:goedel-protocol}
\label[appendix]{app:goedel-bias}

\paragraph{Consensus Protocol.}
To measure post-hoc semantic fidelity (\metricGoedelSM), we adopt the evaluation prompt from Goedel-Prover-V2~\citep{lin2025goedelproverv2}, modifying it only to enforce structured JSON parsing. We deploy \modelGemmaLong{} as the independent evaluator. For each candidate statement, we query the judge four independent times. Each query outputs a categorical verdict of either \emph{Appropriate} or \emph{Not Appropriate}, together with a short free-form rationale. Following \citet{lin2025goedelproverv2}, a formalization is counted as successful under \metricGoedelSM{} if and only if it achieves a unanimous $4/4$ \emph{Appropriate} consensus; we leave this decision rule unchanged.

\paragraph{Prompt bias (from the few-shot examples).}
The demonstrations embedded in the \metricGoedelSM{} prompt show a structural preference for explicit answer injection over open encodings.
\begin{itemize}[nosep, leftmargin=*]
	\item For a problem asking for the intercept of a quadratic polynomial, the prompt marks a formalization \emph{Appropriate} because it injects the resolved numeric value (\texttt{f 3 = 0}).
	\item For a problem asking for the equation of a tangent line, the prompt marks an answer-agnostic formulation \emph{Inappropriate}, on the grounds that \say{\emph{The translation shifts the problem from deriving the tangent line to verifying a given line equation.}}
\end{itemize}
Thus the judge is instructed, by example, to reward closed witnesses and to penalize faithful open formulations that defer the answer.
We do not alter these demonstrations: changing them would change the Goedel protocol itself.

\paragraph{Reject-reason audit.}
To measure how this bias manifests in practice---without perturbing the judge---we attribute Inappropriate votes \emph{post hoc} by inspecting the returned rationales on \datasetOmni{} (answer-agnostic, pass@1).
We first ask an LLM to propose rejection categories from those rationales, consolidate them into a fixed taxonomy of twenty categories, and then ask \modelQwenLong{} to assign every applicable category to each rejection (multi-hot).
\Cref{tab:omni-goedel-reject} reports the three most frequent categories.

\begin{table}[H]
	\centering
	\caption{\textbf{Goedel-SM reject reasons on \datasetOmni{} (Answer-Agnostic) at pass@1.}
		Three most common categories among \emph{Inappropriate} samples (multi-hot; lower is better for defect categories).
		Bold $=$ lowest rate.}
	\label{tab:omni-goedel-reject}
	\small
	\begin{tabular}{@{}l c ccc @{}}
		\toprule
		\textbf{Method}
		& \textbf{$N_{\mathrm{rejection}}$}
		& \textbf{\makecell{Wrong \\ Injection (\%)}}
		& \textbf{\makecell{Mathematically \\ False (\%)}}
		& \textbf{Existential (\%)} \\
		\midrule
		\methodMonolithic
		& 147 & 69.4 & 84.4 & \textbf{8.2} \\
		\methodMonoSemanticLoop
		& 129 & 68.2 & 80.6 & 15.5 \\
		\methodDecomposition
		& 164 & 8.5 & 22.0 & 72.0 \\
		\methodSemanticLoop{} (\methodOurs)
		& 138 & \textbf{7.2} & \textbf{18.1} & 76.8 \\
		\midrule
		\methodGoedel (Finetuned)
		& 204 & 56.9 & 74.5 & 19.6 \\
		\bottomrule
	\end{tabular}
\end{table}

The three reported categories are defined as follows.
\textbf{Wrong Injection:} \SF hard-codes a concrete answer absent from the open \SNL.
\textbf{Mathematically False:} the judge treats \SF as asserting a false claim.
\textbf{Existential:} a faithful open encoding is rejected primarily for omitting an explicit answer.

The empirical split matches the prompt-level bias above.
Monolithic generators and fine-tuned \methodGoedel{} are dominated by Mathematically False and Wrong Injection (\methodGoedel{}: \qty{74.5}{\percent} and \qty{56.9}{\percent} of rejects): they guess a closed witness and often encode an incorrect constant.
Decomposed generators, including \methodOurs{}, are dominated by Existential (\qty{76.8}{\percent}): they keep open encodings ($\exists x,\,P(x)$) that the judge systematically prefers less than explicit answers.
This asymmetry is intentional under our loop contract.
Our in-loop rater is not trained or prompted to verify the mathematical correctness of conjectured constants; doing so would require solving the informal problem and would conflate translation with deduction.
We therefore treat both injection and existential texture as admissible, and we do not route pure mathematical-falsity rejections---especially wrong injected witnesses---through the correction loop.
Accordingly, the residual \metricGoedelSM{} gap should not be read as a failure of decomposition: for \methodGoedel{} and monolithic methods it largely reflects unsuccessful guessing, whereas for \methodOurs{} it largely reflects the judge's documented preference against open existential formalizations.

%% file: appendix_results.tex
\section{Comprehensive Performance Results and Trajectory Analysis}
\label[appendix]{app:full-results}
\label[appendix]{app:trajectory}

We expand the generator-backbone scaling results of \Cref{tab:omni-backend-scaling}
and the human validation of \metricOursSM{} summarized in \Cref{sec:experiments},
reporting the full metric suite, witness-injection rates, and the stratified audit design.

\subsection{Generator Backbone Ablation: Scaling across Model Generations}
\label[appendix]{app:backend-ablation}

To verify that the performance of \methodOurs{} is not contingent upon the specific inductive biases of \modelQwenLong{}, we evaluated our complete decomposed semantic pipeline (\methodOurs{}, $T=5$, pass@1) across three successive model generations: \modelQwenThree{}, \modelQwenLong{}, and \modelQwenThreeEight{}. All runs were conducted on \datasetOmni{} in the Answer-Aware setting and audited using identical protocols: Lean~4 type-checking (\metricCompile), unanimous 4/4 independent \modelGemmaLong{} consensus (\metricGoedelSM), our multi-dimensional semantic gate (\metricOursSM), and our witness injection audit.

\begin{table}[H]
	\centering
	\caption{\textbf{Comprehensive generator backbone ablation on \datasetOmni{} (Answer-Aware, pass@1, $N=\datasetOmniN$).}
		$^*$In-loop optimization target. All methods use \methodOurs{} with identical prompting, repair budgets ($T=5$), and independent judge protocols. Joint solved counts reported out of 300 problems.}
	\label{tab:backend-full-appendix}
	\small
	\begin{tabular}{@{}l ccccc c @{}}
		\toprule
		\textbf{Model Backbone}
		& \textbf{\metricCompile}
		& \textbf{\metricGoedelSM}
		& \textbf{\makecell{\metricCompile $\wedge$ \\ \metricGoedelSM}}
		& \textbf{Joint Count}
		& \textbf{\makecell{\metricCompile $\wedge$ \\ \metricOursSM$^*$}}
		& \textbf{\makecell{Witness \\ Injection (\%)}} \\
		\midrule
		\modelQwenThree{} (32B)       & 57.3 & 46.3 & 30.3 & 91 / 300  & 52.3 & 22.0 \\
		\modelQwenLong{} (27B)        & 96.0 & 80.3 & 77.3 & 232 / 300 & 93.7 & \textbf{90.2} \\
		\modelQwenThreeEight{} (27B)  & \textbf{98.0} & \textbf{86.3} & \textbf{85.7} & \textbf{257 / 300} & \textbf{95.7} & 89.5 \\
		\bottomrule
	\end{tabular}
\end{table}

\paragraph{Analysis of Scaling Trajectories.}
The empirical progression highlights two clear behaviors:
\begin{enumerate}[leftmargin=*, nosep]
	\item \textbf{Monotonic Verification Synergy:} As the generator's underlying formal competence improves, the efficacy of the repair loop compounds dramatically. While \modelQwenThree{} struggles to interpret Lean~4 compiler diagnostics (yielding only \qty{57.3}{\percent} compilation and \qty{30.3}{\percent} joint fidelity), \modelQwenThreeEight{} reaches \textbf{\qty{98.0}{\percent}} compilation and \textbf{\qty{85.7}{\percent}} joint fidelity.
	\item \textbf{Witness Extraction Ceiling:} In Answer-Aware formalization, injecting the target constant from the informal proof is expected. \modelQwenThree{} achieves only \qty{22.0}{\percent} witness injection, indicating frequent comprehension failures on informal mathematical texts. In contrast, both \modelQwenLong{} (\qty{90.2}{\percent}) and \modelQwenThreeEight{} (\qty{89.5}{\percent}) reach the witness extraction ceiling. Consequently, the performance improvement between 3.6 and 3.8 is purely driven by superior formalization fidelity, syntax resolution, and premise alignment.
\end{enumerate}

\begin{revblock}
\subsection{Inference Token Accounting and Leak-Free Fidelity}
\label[appendix]{app:token-accounting}

Because single-shot supervised baselines (\methodGoedelLong{}) and training-free agentic loops allocate compute across different phases, \Cref{tab:token-accounting} reports exact per-problem generation token totals on \datasetOmni{} (Answer-Agnostic, $N=\datasetOmniN$). Reported tokens sum prompt and completion tokens across all active generation stages---including the statement formatter and, for dual-signal configurations, the in-loop \metricOursSM{} rater---while excluding the post-hoc evaluation judge (\metricGoedelSM{}) and Lean compiler invocations. Alongside raw Joint (\metricCompileAndGoedelSM{}), we report \emph{Leak-Free Joint}, counting candidates that compile, pass \metricGoedelSM{}, and do not embed an unverified concrete witness on Question-type queries.

\begin{table}[H]
	\centering
	\caption{\rev{\textbf{Generation token accounting and leak-free joint fidelity on \datasetOmni{} (Answer-Agnostic, $N=\datasetOmniN$).} Tokens average prompt plus completion tokens across the generation pipeline (including formatting and, for dual-signal variants, the in-loop rater; post-hoc scoring judge excluded). $T$ denotes the maximum repair iteration cap.}}
	\label{tab:token-accounting}
	\small
	\resizebox{\textwidth}{!}{%
	\color{revblue}
	\begin{tabular}{@{}l l r rrr@{}}
		\toprule
		\textbf{Method} & \textbf{Training} & \textbf{Avg.\ Tokens} & \textbf{Joint (\%)} & \textbf{Leak-Free Joint (\%)} & \textbf{Leakage ($\downarrow$, \%)} \\
		\midrule
		\methodMonolithic{} (pass@1, $T=1$) & None & 6.3k & 40.0 & 11.3 & 72.6 \\
		\textsf{Monolithic + Anti-Solver} (pass@1, $T=1$) & None & 6.8k & 34.7 & 14.3 & 63.0 \\
		\methodMonoSyntaxLoop{} (pass@1, $T=5$) & None & 12.2k & 49.0 & 13.7 & 76.4 \\
		\methodMonoSemanticLoop{} (pass@1, $T=5$) & None & 37.3k & 55.3 & 18.0 & 70.9 \\
		\methodDecomposition{} (pass@1, $T=1$) & None & 18.5k & 29.7 & 29.7 & \textbf{1.0} \\
		\methodSyntaxLoop{} (pass@1, $T=5$) & None & 25.5k & 42.7 & 42.0 & 2.7 \\
		\textbf{\methodOurs{}} (pass@1, $T=5$) & None & 63.8k & \textbf{50.3} & \textbf{48.7} & 2.7 \\
		\methodGoedelLong{} (pass@1) & Full SFT (32B) & 1.3k & 26.7 & 13.3 & 56.2 \\
		\midrule
		\methodMonolithic{} (pass@4, $T=1$) & None & 25.3k & 59.0 & 16.7 & 73.3 \\
		\textbf{\methodOurs{}} (pass@4, $T=5$) & None & 252k & \textbf{73.3} & \textbf{69.3} & \textbf{4.7} \\
		\methodGoedelLong{} (pass@4) & Full SFT (32B) & 5.3k & 42.0 & 20.0 & 58.1 \\
		\bottomrule
	\end{tabular}%
	}
\end{table}

\Cref{tab:token-split} decomposes pass@1 token usage into the initial draft generation phase (compounding generator stages and statement formatting), the iterative repair phase (combining corrector and, where active, in-loop \metricOursSM{} rater tokens), and the average number of executed repair loops per problem. Crucially, using fewer than half the tokens of \methodMonoSemanticLoop{} (\num{18.5}k vs.\ \num{37.3}k) with zero repair rounds, single-pass \methodDecomposition{} already achieves $+11.7$ percentage points higher Leak-Free Joint (\qty{29.7}{\percent} vs.\ \qty{18.0}{\percent}), and single-attempt \methodOurs{} (\qty{48.7}{\percent}) more than doubles four-sample \methodGoedelLong{} (\qty{20.0}{\percent}).

\begin{table}[H]
	\centering
	\caption{\rev{\textbf{Phase-level pass@1 token breakdown and repair loop counts on \datasetOmni{} (Answer-Agnostic, $N=\datasetOmniN$).} Draft Phase compounds initial synthesis and statement formatting; Repair Phase sums all iterative correction and in-loop rater tokens across active repair rounds.}}
	\label{tab:token-split}
	\small
	{\color{revblue}
	\begin{tabular}{@{}l rrrr@{}}
		\toprule
		\textbf{Method} & \textbf{Draft Phase} & \textbf{Repair Phase} & \textbf{Avg.\ Repair Loops} & \textbf{Total Tokens} \\
		\midrule
		\methodGoedelLong{} & 1.3k & 0 & 0 & 1.3k \\
		\methodMonolithic{} ($T=1$) & 6.3k & 0 & 0 & 6.3k \\
		\textsf{Monolithic + Anti-Solver} ($T=1$) & 6.8k & 0 & 0 & 6.8k \\
		\methodMonoSyntaxLoop{} ($T=5$) & 6.3k & 5.9k & 0.54 & 12.2k \\
		\methodMonoSemanticLoop{} ($T=5$) & 6.3k & 31.0k & 0.91 & 37.3k \\
		\methodDecomposition{} ($T=1$) & 18.5k & 0 & 0 & 18.5k \\
		\methodSyntaxLoop{} ($T=5$) & 18.5k & 7.0k & 0.69 & 25.5k \\
		\textbf{\methodOurs{}} ($T=5$) & 18.5k & 45.3k & 1.49 & 63.8k \\
		\bottomrule
	\end{tabular}%
	}
\end{table}

\subsection{Prompt-Only Anti-Solver Control vs.\ Architectural Stage Isolation}
\label[appendix]{app:anti-solver}

To test whether monolithic answer leakage stems merely from omitting an instruction forbidding answer computation, we evaluated a controlled \textsf{Monolithic + Anti-Solver} variant on \modelQwenLong{} ($T=1$, pass@1, \num{6.8}k avg.\ tokens) across all three benchmarks. The prompt enforces strict no-calculation, no-witness-injection, and single-\lean{sorry} rules directly within the monolithic formalizer while omitting the Distiller and Preprocessor stages. As shown in \Cref{tab:anti-solver}, prompt-level instructions alone reduce leakage on \datasetOmni{} only marginally (from \qty{71.7}{\percent} to \qty{61.0}{\percent} across all $N{=}300$ drafts, and from \qty{72.6}{\percent} to \qty{63.0}{\percent} among compiling Question-type drafts in \Cref{tab:token-accounting}), lifting Leak-Free Joint only from \qty{11.3}{\percent} to \qty{14.3}{\percent} while degrading raw Joint fidelity from \qty{40.0}{\percent} to \qty{34.7}{\percent} (and from \qty{20.0}{\percent} to \qty{10.9}{\percent} on \datasetImo{}). By contrast, \methodDecomposition{} ($T=1$) collapses leakage to \qty{1.0}{\percent} (\qty{29.7}{\percent} Leak-Free Joint) without repair, demonstrating that architectural stage isolation---separating declarative goal reformulation in the Distiller from Lean syntax synthesis---is essential to prevent answer guessing.

\begin{table}[H]
	\centering
	\caption{\rev{\textbf{Prompt-only Anti-Solver control on \methodMonolithic{} (Answer-Agnostic, pass@1).} Comparing standard \methodMonolithic{} prompting against an explicit no-solving / no-injection / single-\lean{sorry} monolithic prompt on \modelQwenLong{} ($T=1$; \textbf{Leak} reported across all $N$ drafts).}}
	\label{tab:anti-solver}
	\small
	{\color{revblue}
	\begin{tabular}{@{}l r ccc ccc@{}}
		\toprule
		& & \multicolumn{3}{c}{\textbf{Standard \methodMonolithic{}}} & \multicolumn{3}{c}{\textbf{\methodMonolithic{} + Anti-Solver}} \\
		\cmidrule(lr){3-5} \cmidrule(l){6-8}
		\textbf{Dataset} & \textbf{$N$} & \textbf{\metricCompile} & \textbf{Joint} & \textbf{Leak ($\downarrow$)} & \textbf{\metricCompile} & \textbf{Joint} & \textbf{Leak ($\downarrow$)} \\
		\midrule
		\datasetOmni{} & 300 & 73.0 & 40.0 & 71.7 & 73.0 & 34.7 & 61.0 \\
		\datasetImo{} & 175 & 58.3 & 20.0 & 21.7 & 55.4 & 10.9 & 13.1 \\
		\datasetImoFormalized{} & 223 & 72.6 & 43.0 & 31.8 & 74.4 & 41.3 & 23.8 \\
		\bottomrule
	\end{tabular}%
	}
\end{table}
\end{revblock}

\section{Human Expert Validation of the Semantic Gate}
\label[appendix]{app:human-validation}

To evaluate the reliability of our automated quality gate, three Lean-proficient annotators independently scored an adversarial audit corpus of $N=45$ zero-shot \datasetOmni{} formalizations under the same five-core rubric and binary gate used by \metricOursSM{}.
We compare the automated gate to \textbf{majority vote}.

\paragraph{Stratified Audit Design.}
Rather than drawing a naive uniform random sample (which would be dominated by straightforward instances), the benchmark was deliberately stratified across four distinct categories to stress-test the gate against edge cases:
\begin{enumerate}[leftmargin=*, nosep]
	\item \textbf{Silent Semantic Failures ($N=25$):} Statements that compile cleanly in Lean~4 but contain subtle mathematical flaws (e.g., incorrect summation indices, artificial domain restrictions, or hardcoded constants).
	\item \textbf{Compile-Failed Faithful Drafts ($N=10$):} Mathematically sound translations that failed type-checking solely due to minor syntactic or import errors.
	\item \textbf{Verified Clean Translations ($N=5$):} Idiomatic, fully compilable, and semantically equivalent formalizations.
	\item \textbf{Catastrophic Failures ($N=5$):} Formalizations that are both uncompilable and mathematically false.
\end{enumerate}

\paragraph{Results.}
Across the stratified audit, the automated gate concurs with majority human judgment on \qty{71.1}{\percent} of formalizations.
Beyond the binary decision, mean human Semantic Match aligns with the rater at Pearson $r = 0.64$ and Spearman $\rho = 0.63$, providing complementary evidence that the evaluator ranks drafts consistently with human semantic preference.

\begin{revblock}
\subsection{Blinded Expert Head-to-Head Comparison on Final IMO Outputs}
\label[appendix]{app:blind-imo-human}

Whereas the $N=45$ stratified audit above stress-tests our internal gate on failure-enriched drafts, we additionally conducted a blinded human head-to-head evaluation comparing final pass@1 outputs from \methodOurs{} and \methodGoedelLong{} on randomly sampled problems.

\paragraph{Sampling and Blinding Protocol.}
To ensure both systems target identical mathematical claims without regime-revealing packaging differences (existential queries vs.\ injected answers), we restricted sampling to assertion-type problems in \datasetImoFormalized{} ($111$ assertion problems out of $223$, of which $80$ compile under pass@1 for both systems) and drew $N=30$ pairs uniformly at random. Before annotation, surface fingerprints were stripped from every file (removing \methodGoedel{}'s boilerplate \lean{import Aesop}, \lean{set\_option maxHeartbeats 0}, and \lean{open BigOperators Real Nat Topology Rat} headers, stripping informal comments, and renaming all target declarations to \lean{theorem statement}), and system presentation order (System~A vs.\ System~B) was randomized independently per problem. Three Lean-proficient annotators independently judged each pair strictly for mathematical fidelity to the informal contest statement (hypotheses, quantifiers, conclusion, and non-triviality), totaling $30 \times 3 = 90$ ratings.

\paragraph{Results.}
\Cref{tab:blind-imo-human} summarizes the unblinded panel outcomes (substantial pairwise agreement between conservative raters, Cohen's $\kappa = 0.72$). By majority vote ($\ge 2/3$ annotators), experts preferred \methodOurs{} on $12/30$ problems (\qty{40.0}{\percent}, with $7$ unanimous $3/3$ wins), judged $17/30$ (\qty{56.7}{\percent}) as Ties where both formalizations were faithful, and preferred \methodGoedelLong{} on $0/30$ (\qty{0.0}{\percent}), with $1$ evenly split $1$--$1$--$1$ vote---yielding a $12$--$0$ margin on decisive majorities.

\begin{table}[H]
	\centering
	\caption{\rev{\textbf{Blinded three-expert preference audit on $N=30$ randomly sampled compiling assertion-type \datasetImoFormalized{} problems} (\methodOurs{} vs.\ \methodGoedelLong{}, pass@1, $90$ expert ratings).}}
	\label{tab:blind-imo-human}
	\small
	{\color{revblue}
	\begin{tabular}{@{}l r r@{}}
		\toprule
		\textbf{3-Expert Panel Outcome} & \textbf{Count ($N=30$)} & \textbf{Share (\%)} \\
		\midrule
		\textbf{\methodOurs{} Preferred} ($\ge 2/3$ majority; $7$ unanimous) & \textbf{12} & \textbf{40.0} \\
		Tie ($\ge 2/3$ majority; both faithful) & 17 & 56.7 \\
		\methodGoedelLong{} Preferred ($\ge 2/3$ majority) & 0 & 0.0 \\
		No Majority ($1$--$1$--$1$ split) & 1 & 3.3 \\
		\bottomrule
	\end{tabular}%
	}
\end{table}

\subsection{Lean 4 Kernel Equivalence Audit on IMO Assertions}
\label[appendix]{app:beq-first40}

To complement LLM-based semantic scoring and blinded human preferences with machine-checked logical verification, we evaluated bidirectional semantic equivalence (\textbf{BEq}) across all $91$ compiling, \metricGoedelSM{}-passing assertion problems in \datasetImoFormalized{} (spanning \ac{IMO} 1959--2024).

\paragraph{Kernel Equivalence Protocol.}
For each problem pair, both the human reference statement from \textsf{Compfiles} and the \methodOurs{} pass@1 formalization are loaded verbatim into isolated namespaces (\lean{BeqGold} and \lean{BeqGen}) with their source proofs replaced by \lean{:= by sorry}. We then query an LLM with iterative Lean~4 compiler feedback to synthesize a formal bidirectional bridge:
\[
\texttt{statements\_equivalent} : \SF^{\text{Gold}} \longleftrightarrow \SF^{\methodOurs}.
\]
The compiled Lean~4 kernel environment is inspected directly and certifies equivalence \emph{if and only if}: (i)~the theorem type matches the exact biconditional between the unaltered \textsf{Compfiles} and \methodOurs{} propositions; (ii)~no helper definitions are rewritten, weakened, or strengthened; (iii)~recursive transitive axiom inspection via \lean{Lean.collectAxioms} restricts dependencies strictly to standard foundational Lean~4 axioms ($\{\texttt{propext}, \texttt{Classical.choice}, \texttt{Quot.sound}\}$), rejecting any occurrence of \lean{sorryAx} (which prevents a bridge proof from invoking either unproved theorem); and (iv)~an LLM auditor verifies that the proof genuinely transports the target hypothesis rather than cheating by proving both theorems independently from scratch ($\texttt{True} \leftrightarrow \texttt{True}$).

\begin{table}[H]
	\centering
	\caption{\rev{\textbf{Lean~4 kernel equivalence evaluation across all $91$ compiling \datasetImoFormalized{} assertions} (\ac{IMO} 1959--2024; bridges synthesized with an LLM and iterative compiler feedback).}}
	\label{tab:beq-first40}
	\small
	{\color{revblue}
	\begin{tabular}{@{} >{\raggedright\arraybackslash}p{0.42\textwidth} r r >{\raggedright\arraybackslash}p{0.31\textwidth} @{}}
		\toprule
		\textbf{Outcome Category} & \textbf{Count} & \textbf{Share (\%)} & \textbf{Verification Basis} \\
		\midrule
		\textbf{Certified Kernel Bridges} & \textbf{70} & \textbf{76.9} & Sorry-free $\SF^{\text{Gold}} \leftrightarrow \SF^{\methodOurs}$ accepted by Lean~4 \\
		\addlinespace[3pt]
		\textbf{\textsf{Compfiles} Reference Defects / Incomplete} & \textbf{8} & \textbf{8.8} & Human gold vacuous, altered, or ships open \texttt{sorry} \\
		\hspace{1em}\emph{--- \methodOurs{} strictly more faithful than Gold} & \emph{4} & \emph{4.4} & Imo1972P1, Imo1972P5, Imo2017P6, Imo2022P3 \\
		\hspace{1em}\emph{--- Reference alters hypothesis / open \texttt{sorry}} & \emph{4} & \emph{4.4} & Imo1969P2, Imo1987P1, Imo2001P3, Imo2024P3 \\
		\addlinespace[3pt]
		\textbf{Unresolved Domain / Representation Differences} & \textbf{11} & \textbf{12.1} & Different Mathlib domain/encoding; neither BEq nor $\neg$BEq proven \\
		\addlinespace[3pt]
		\textbf{Degenerate Proof Caught by LLM Verification} & \textbf{1} & \textbf{1.1} & Imo1964P2 (both sides proved from scratch; excluded) \\
		\addlinespace[3pt]
		\textbf{Unilateral \methodOurs{} Formalization Slip} & \textbf{1} & \textbf{1.1} & Imo1967P3 (truncating \texttt{Nat} subtraction) \\
		\midrule
		\textbf{Total Evaluated Assertions (\ac{IMO} 1959--2024)} & \textbf{91} & \textbf{100.0} & All compiling, \metricGoedelSM{}-passing assertions \\
		\bottomrule
	\end{tabular}%
	}
\end{table}

\paragraph{Results and Forensic Breakdown of Non-Bridged Pairs.}
As reported in \Cref{tab:beq-first40}, the Lean~4 kernel formally certifies bidirectional logical equivalence on \textbf{$70/91$ (\qty{76.9}{\percent})} problems. Our LLM anti-cheating verification caught exactly \textbf{$1/91$ (\qty{1.1}{\percent})} degenerate kernel pass (Imo1964P2, where the prover closed $\SF^{\text{Gold}} \leftrightarrow \SF^{\methodOurs}$ by proving both triangle inequalities independently from scratch and discarding the bridge hypothesis; we exclude this from certified bridges). Among the remaining non-bridged pairs, only \textbf{$1/91$ (\qty{1.1}{\percent})} reflects a genuine unilateral \methodOurs{} translation error (Imo1967P3, where natural-number subtraction \texttt{m - 1 : Nat} truncates at zero instead of encoding integer divisibility). For \textbf{$11/91$ (\qty{12.1}{\percent})} problems, we could not establish a formal bridge because the human reference and \methodOurs{} formalize the problem over different domains or Mathlib representations---such as \texttt{EuclideanSpace R (Fin 2)} vs.\ $\mathbb{R} \times \mathbb{R}$ coordinates (Imo1970P6), root distinctness via \texttt{Multiset.Nodup} vs.\ non-vanishing polynomial discriminant \texttt{Polynomial.discr P != 0} (Imo1976P2), and abstract metric triangles vs.\ algebraic side-length inequalities (Imo1983P6)---leaving neither formal equivalence (BEq) nor inequivalence ($\neg$BEq) proven.

The remaining \textbf{$8/91$ (\qty{8.8}{\percent})} discrepancies stem from verified defects or unfinished \lean{sorry} stubs in the human \textsf{Compfiles} reference statements themselves:
\begin{enumerate}[leftmargin=*, itemsep=3pt, topsep=4pt]
	\item \textbf{Imo1972P1 (Gold Vacuous; \methodOurs{} Strictly More Faithful):} From any $10$-element set of two-digit numbers, the contest asks to select two disjoint subsets with equal sum. \textsf{Compfiles} omits subset non-emptiness (\texttt{exists S1 S2 : Finset Nat, S1 <= S /\ S2 <= S /\ Disjoint S1 S2 /\ ...}), making the reference theorem vacuously provable for \emph{any} set $S$ by instantiating $S_1 = S_2 = \emptyset$ (Case Study~3 in \Cref{app:imo-case-studies}). \methodOurs{} explicitly enforces \texttt{S1.Nonempty /\ S2.Nonempty}, faithfully preserving the pigeonhole problem.
	\item \textbf{Imo1972P5 (Gold Over-Generalized; \methodOurs{} Strictly More Faithful):} Given $f(x+y) + f(x-y) = 2f(x)g(y)$ with $f$ not identically zero and $|f(x)| \le 1$ for all $x \in \mathbb{R}$, the contest asks to prove $|g(y)| \le 1$. \textsf{Compfiles} weakens the hypothesis $|f(x)| \le 1$ to boundedness above, \texttt{(hf2 : BddAbove (Set.range fun x => ||f x||))} (which holds, e.g., for $f = 2\cos,\, g = \cos$ where $\|f\|_\infty = 2 > 1$). \methodOurs{} preserves the exact contest hypothesis $\forall x,\, |f(x)| \le 1$.
	\item \textbf{Imo2017P6 (Gold False via Off-by-One Sum Range; \methodOurs{} Strictly More Faithful):} The contest requires a homogeneous polynomial $\sum_{i=0}^{n} c_i x^i y^{n-i} = 1$ of degree $n$ over all $(x,y) \in S$. \textsf{Compfiles} sums over \texttt{Finset.range n} ($i \in \{0,\dots,n-1\}$), dropping the $x^n$ term; on $S = \{(1,0)\}$, every term has a positive power of $y=0$ and the theorem collapses to $0 = 1$ (formally refuted in Lean~4). \methodOurs{} sums over all $n+1$ terms via \texttt{Fin (n + 1)}.
	\item \textbf{Imo2022P3 (Gold False at Empty Set; \methodOurs{} Strictly More Faithful):} The contest quantifies over finite sets $S$ of odd primes without assuming $S \neq \emptyset$. \textsf{Compfiles} concludes $\exists i : \texttt{Fin \#S},\, \dots$; at $S = \emptyset$, $\texttt{Fin 0}$ is uninhabited, rendering the conclusion false while all hypotheses hold vacuously (formally refuted in Lean~4). \methodOurs{} formalizes circular arrangements via inhabited list placements.
	\item \textbf{Imo1969P2 \& Imo2001P3 (Both Imperfect; Gold Injects Hypothesis or Contradicts Docstring):} In Imo1969P2, \textsf{Compfiles} injects an unstated condition $f(0) \neq 0$ (\texttt{h4 : sum i in Finset.range n, (Real.cos (a i) / (2 \^{} i)) != 0}), excluding valid contest inputs (whereas \methodOurs{} omits $n \ge 1$). In Imo2001P3, \textsf{Compfiles}'s natural-language docstring states that each girl--boy pair shares \emph{at most} one problem, whereas its Lean code writes \texttt{\ensuremath{\lnot}Disjoint} (\emph{at least} one); fed the docstring, \methodOurs{} faithfully encodes \texttt{card <= 1} (which is separately refutable on empty solve sets).
	\item \textbf{Imo1987P1 \& Imo2024P3 (Gold Ships Open \texttt{sorry} in Statement / Type):} In Imo1987P1, \textsf{Compfiles} leaves finiteness of the fixed-point fiber as \texttt{Fintype (fiber alpha k) := by sorry}, whereas \methodOurs{} defines fixed-point counts directly; in Imo2024P3, \textsf{Compfiles} leaves \texttt{def EventuallyPeriodic ... := sorry} inside the theorem type, whereas \methodOurs{} writes out periodicity explicitly.
\end{enumerate}
\end{revblock}

\section{External Evaluation of the Semantic Gate}
\label[appendix]{app:gate-external-calibration}

\Cref{app:human-validation} validates the gate on an adversarial expert audit.
Complementing that study, we evaluate the same binary gate on two public autoformalization-alignment benchmarks with gold faithfulness labels.

\paragraph{Benchmarks.}
ConsistencyCheck~\citep{chen2026reform} comprises $N{=}859$ informal--formal pairs drawn from miniF2F and ProofNet sources ($570$ labeled faithful, $289$ unfaithful).
ProofNetVerif~\citep{poiroux2025reliable} comprises $N{=}3752$ pairs in which a candidate Lean~4 formalization is scored against an informal statement ($1142$ labeled correct, $2610$ incorrect).
Each item provides a natural-language statement, a Lean candidate, and a binary gold label indicating semantic faithfulness.

\paragraph{Protocol.}
For every informal--formal pair we run the in-loop semantic rater zero-shot and apply the identical binary gate used at test time, which returns a single Pass or Fail decision.
Pass is treated as the positive prediction and compared to the benchmark gold label.
No decision rule is retuned on these corpora.

\paragraph{Metrics.}
\Cref{tab:gate-external-calibration} reports:
\begin{itemize}[nosep,leftmargin=*]
	\item \textbf{Accuracy:} fraction of pairs on which the gate matches the gold label;
	\item \textbf{Precision / Recall / F1:} for the Pass class relative to gold-positive items;
	\item \textbf{Matthews correlation:} chance-adjusted association between predictions and gold labels, ranging in $[-1,1]$ (higher is better; $0$ is chance-level);
	\item \textbf{Majority baseline:} accuracy obtained by always predicting the more frequent gold class.
\end{itemize}

\paragraph{Results.}
On both benchmarks the gate exceeds the majority baseline by about $19$ percentage points, reaching \qty{85.7}{\percent} accuracy on ConsistencyCheck and \qty{88.8}{\percent} on ProofNetVerif, with Matthews correlations of $0.67$ and $0.74$ respectively (\Cref{tab:gate-external-calibration}).

\begin{table}[H]
	\centering
	\caption{\textbf{External evaluation of the \metricOursSM{} five-core gate}
		against gold faithfulness labels on ConsistencyCheck and ProofNetVerif.
		Predictor: published Pass/Fail gate; Pass is the positive class.}
	\label{tab:gate-external-calibration}
	\resizebox{\textwidth}{!}{%
	\begin{tabular}{@{}l r ccccc c @{}}
		\toprule
		\textbf{Benchmark}
		& \textbf{$N$}
		& \textbf{Accuracy (\%)}
		& \textbf{Precision (\%)}
		& \textbf{Recall (\%)}
		& \textbf{F1 (\%)}
		& \textbf{\makecell{Matthews\\correlation}}
		& \textbf{\makecell{Majority\\baseline (\%)}} \\
		\midrule
		ConsistencyCheck~\citep{chen2026reform}
		& 859 & \textbf{85.7} & 88.0 & 90.9 & 89.4 & \textbf{0.67} & 66.4 \\
		ProofNetVerif~\citep{poiroux2025reliable}
		& 3752 & \textbf{88.8} & 82.1 & 80.9 & 81.5 & \textbf{0.74} & 69.6 \\
		\bottomrule
	\end{tabular}%
	}
\end{table}

\section{Blind Pairwise Preferences on \datasetImo{}}
\label[appendix]{app:imo-pairwise}

\Cref{tab:imo-pairwise} reports the full order-swapped Answer-Agnostic pass@1 preference breakdown summarized in \Cref{sec:experiments}: an independent automated judge compares \methodOurs{} against \methodGoedel{} on the unformalized IMO frontier.

\begin{table}[H]
	\centering
	\caption{\textbf{Blind pairwise preferences on \datasetImo{} (Answer-Agnostic, pass@1).} Percentages may not sum to \qty{100}{\percent} due to rounding.}
	\label{tab:imo-pairwise}
	\small
	\begin{tabular}{@{}l ccccc @{}}
		\toprule
		& \multicolumn{5}{c}{\textbf{Preference for \methodOurs{} over \methodGoedel (\%)}} \\
		\cmidrule(l){2-6}
		\textbf{Evaluation Type} & \textbf{Much Better} & \textbf{Better} & \textbf{Tie} & \textbf{Worse} & \textbf{Much Worse} \\
		\midrule
		Semantic (\metricPWR) & 78.3 & 1.1 & 17.7 & 1.1 & 1.7 \\
		\bottomrule
	\end{tabular}
\end{table}

\begin{revblock}
\subsection{Expert-Verified IMO Case Studies}
\label[appendix]{app:imo-case-studies}

To complement aggregate metrics on \datasetImo{} and \datasetImoFormalized{}, we present three expert-audited Lean~4 formalizations generated by \methodOurs{}: two Euclidean and solid geometry problems from \datasetImo{} that had never previously been formalized in Lean~4, followed by a combinatorics problem from \datasetImoFormalized{} where the community \textsf{Compfiles} reference contains a vacuity bug that \methodOurs{} avoids.

\medskip
\noindent\begin{minipage}{\linewidth}
\noindent\textbf{Case Study 1 (\datasetImo{} --- Never Previously Formalized): IMO 1966, Problem 6.}\par\nopagebreak\vspace{4pt}
\noindent\emph{Informal statement:} ``In the interior of sides $BC, CA, AB$ of triangle $ABC$, any points $K, L, M$, respectively, are selected. Prove that the area of at least one of the triangles $AML, BKM, CLK$ is less than or equal to one quarter of the area of triangle $ABC$.''\par\vspace{4pt}
\noindent\methodOurs{} explicitly models signed-determinant planar area, strict open-segment side interiors ($0 < t < 1$), and triangle non-degeneracy (\lean{Area A B C > 0}) in a self-contained single-\lean{sorry} declaration:
\begin{lstlisting}[style=revlean, aboveskip=6pt, belowskip=6pt]
import Mathlib

noncomputable section

def Area (A B C : ℝ × ℝ) : ℝ :=
  (1/2) * |(B.1 - A.1) * (C.2 - A.2) - (B.2 - A.2) * (C.1 - A.1)|

def InteriorOnSide (P : ℝ × ℝ) (A B : ℝ × ℝ) : Prop :=
  ∃ t : ℝ, 0 < t ∧ t < 1 ∧ P = (1 - t) • A + t • B

def NonDegenerateTriangle (A B C : ℝ × ℝ) : Prop :=
  Area A B C > 0

theorem triangle_area_min_bound
  (A B C K L M : ℝ × ℝ)
  (h_nondeg : NonDegenerateTriangle A B C)
  (hK : InteriorOnSide K B C)
  (hL : InteriorOnSide L C A)
  (hM : InteriorOnSide M A B) :
  min (Area A M L) (min (Area B K M) (Area C L K)) ≤ (1/4) * Area A B C := by sorry
\end{lstlisting}
\end{minipage}\par

\bigskip
\noindent\begin{minipage}{\linewidth}
\noindent\textbf{Case Study 2 (\datasetImo{} --- Never Previously Formalized): IMO 1968, Problem 4.}\par\nopagebreak\vspace{4pt}
\noindent\emph{Informal statement:} ``Prove that in every tetrahedron there is a vertex such that the three edges meeting there have lengths which are the sides of a triangle.''\par\vspace{4pt}
\noindent\methodOurs{} encodes a non-degenerate three-dimensional tetrahedron via \lean{AffineIndependent $\mathbb{R}$ v} in \lean{EuclideanSpace $\mathbb{R}$ (Fin 3)}, indexes the three incident edges at each vertex without self-loops, and enforces all three strict triangle inequalities:
\begin{lstlisting}[style=revlean, aboveskip=6pt, belowskip=6pt]
import Mathlib

noncomputable section

structure Tetrahedron where
  v : Fin 4 → EuclideanSpace ℝ (Fin 3)
  non_degenerate : AffineIndependent ℝ v

def otherVertex (i : Fin 4) (j : Fin 3) : Fin 4 :=
  let n := (j : ℕ)
  let m := (i : ℕ)
  if n < m then Fin.mk n (by omega) else Fin.mk (n + 1) (by omega)

def incidentEdgeLengths (T : Tetrahedron) (i : Fin 4) : Fin 3 → ℝ :=
  fun j => dist (T.v i) (T.v (otherVertex i j))

def satisfiesStrictTriangleInequalities (a b c : ℝ) : Prop :=
  a + b > c ∧ a + c > b ∧ b + c > a

theorem tetrahedron_vertex_triangle_inequality (T : Tetrahedron) :
  ∃ i : Fin 4, satisfiesStrictTriangleInequalities
    (incidentEdgeLengths T i 0)
    (incidentEdgeLengths T i 1)
    (incidentEdgeLengths T i 2) := by sorry
\end{lstlisting}
\end{minipage}\par

\bigskip
\noindent\textbf{Case Study 3 (\datasetImoFormalized{} --- Human Reference Vacuity Repaired by \methodOurs{}): IMO 1972, Problem 1.}\par\nopagebreak\vspace{4pt}
\noindent\emph{Informal statement:} ``Prove that from a set of ten distinct two-digit numbers (in decimal), it is possible to select two disjoint subsets whose members have the same sum.''\par\vspace{4pt}
\noindent In the community \textsf{Compfiles} benchmark (Imo1972P1), the human formalization omits subset non-emptiness, allowing $S_1 = \emptyset$ and $S_2 = \emptyset$ to satisfy \lean{Disjoint $\emptyset$ $\emptyset$} and $0 = 0$ vacuously for \emph{any} set $S$ regardless of cardinality or digit constraints:\par\vspace{4pt}
\noindent\begin{minipage}{\linewidth}
\begin{lstlisting}[style=revlean, aboveskip=2pt, belowskip=2pt]
-- Human reference in Compfiles (Imo1972P1; vacuously true via S1 = ∅, S2 = ∅):
theorem imo1972_p1 (S : Finset ℕ)
    (Scard : S.card = 10)
    (Sdigits : ∀ n ∈ S, (Nat.digits 10 n).length = 2) :
    ∃ S1 S2 : Finset ℕ, S1 ⊆ S ∧ S2 ⊆ S ∧
       Disjoint S1 S2 ∧ ∑ n ∈ S1, n = ∑ n ∈ S2, n := by sorry
\end{lstlisting}
\end{minipage}\par\vspace{6pt}
\noindent By contrast, \methodOurs{}'s Preprocessor and in-loop semantic gate explicitly surface and enforce subset non-emptiness (\lean{S1.Nonempty $\wedge$ S2.Nonempty}), blocking the $\emptyset = \emptyset$ trivialization and preserving the intended pigeonhole problem:\par\vspace{4pt}
\noindent\begin{minipage}{\linewidth}
\begin{lstlisting}[style=revlean, aboveskip=2pt, belowskip=2pt]
-- Sage formalization (enforces non-empty disjoint subsets):
import Mathlib

namespace Imo1972P1

theorem imo1972_p1 (S : Finset ℕ)
    (Scard : S.card = 10)
    (Sdigits : ∀ n ∈ S, (Nat.digits 10 n).length = 2) :
    ∃ S1 S2 : Finset ℕ, S1 ⊆ S ∧ S2 ⊆ S ∧
       Disjoint S1 S2 ∧ S1.Nonempty ∧ S2.Nonempty ∧ ∑ n ∈ S1, n = ∑ n ∈ S2, n := by sorry

end Imo1972P1
\end{lstlisting}
\end{minipage}\par
\end{revblock}